%% file: main.tex
\documentclass[10pt,twocolumn,letterpaper]{article}

\usepackage{style}
\date{}
\usepackage[pagebackref,breaklinks,colorlinks]{hyperref}

\usepackage{multirow}
\usepackage{array}
\usepackage{arydshln}
\usepackage{subcaption}
\usepackage{amsmath}

\usepackage{subcaption}
\usepackage{multirow}
\usepackage{arydshln}

\usepackage{adjustbox}

\newcommand{\tablesize}{\fontsize{9}{10}\selectfont}

\newcommand{\smallfbox}[1]{%
  \setlength{\fboxsep}{1pt}
  \setlength{\fboxrule}{0.5pt}
  \fbox{#1}
}

\title{One-Frame Calibration with Siamese Network in Facial Action Unit Recognition}

\author{Shuangquan Feng \hspace{1cm} Virginia R. de Sa\\
University of California San Diego, La Jolla, CA 92093\\
{\tt\small \{s1feng,desa\}@ucsd.edu}
}

\begin{document}
\maketitle
\input{sec/main}
{
    \small
    \bibliographystyle{ieeenat_fullname}
    \bibliography{main}
}
\input{sec/appendix}

\end{document}

%% file: sec/main.tex
\begin{abstract}
Automatic facial action unit (AU) recognition is used widely in facial expression analysis. Most existing AU recognition systems aim for cross-participant non-calibrated generalization (NCG) to unseen faces without further calibration. However, due to the diversity of facial attributes across different identities, accurately inferring AU activation from single images of an unseen face is sometimes infeasible, even for human experts---it is crucial to first understand how the face appears in its neutral expression, or significant bias may be incurred. Therefore, we propose to perform one-frame calibration (OFC) in AU recognition: for each face, a single image of its neutral expression is used as the reference image for calibration. With this strategy, we develop a Calibrating Siamese Network (CSN) for AU recognition and demonstrate its remarkable effectiveness with a simple iResNet-50 (IR50) backbone. On the DISFA, DISFA+, and UNBC-McMaster datasets, we show that our OFC CSN-IR50 model (a) substantially improves the performance of IR50 by mitigating facial attribute biases (including biases due to wrinkles, eyebrow positions, facial hair, etc.), (b) substantially outperforms the naive OFC method of baseline subtraction as well as (c) a fine-tuned version of this naive OFC method, and (d) also outperforms state-of-the-art NCG models for both AU intensity estimation and AU detection. The code is available at https://github.com/ShuangquanFeng/CSN.
\end{abstract}

\section{Introduction}

\begin{figure*}[htb]
    \centering
    \includegraphics[width=0.95\linewidth]{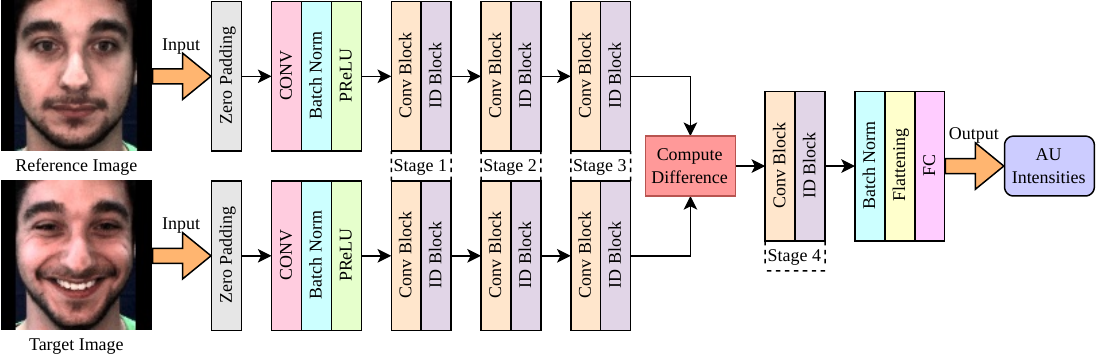}
    \caption{The CSN-IR50 network architecture. The reference image and the target image are fed into two identical networks with shared weights and joined in an intermediate stage of the network by computing the difference between their feature maps. The example reference image and target image are from the DISFA dataset.}
    \label{fig:CSN-IR50}
\end{figure*}

Facial expression analysis is important for understanding human emotions and behaviors across various fields, including human-computer interaction, psychology, and security.
The Facial Action Coding System (FACS) is a comprehensive system for describing human facial movement developed by Ekman and Friesen \cite{ekman1978facial}, widely recognized and extensively used in facial expression analysis for its ability to describe facial movement objectively and systematically. It breaks down facial expressions into individual components of muscle movement, called action units (AUs). 

As manual AU coding is expensive and time-consuming, automatic AU recognition is used widely in facial expression analysis. Most existing AU recognition systems aim for cross-participant non-calibrated generalization (NCG) to unseen faces \cite{baltruvsaitis2016openface,zhao2016deep,shao2018deep,fan2020facial,liu2023multi,yuan2024auformer}, where individual images/frames of the faces are fed into the trained model as input for AU recognition. However, due to the diversity of facial attributes across different identities, accurately inferring AU activation from single images of an unseen face is sometimes infeasible, even for human experts---it is crucial to first understand how the face appears in its neutral expression, or significant bias may be incurred. In the official FACS manual \cite{ekman2002facial}, the importance of taking the face's neutral appearance into account as the baseline is repeatedly emphasized for human scoring of various AUs. Firstly, without understanding the neutral appearance, permanent facial features (e.g. wrinkles, bulges, pouches) may be misidentified as evidence for AU activation. Secondly, scoring of many AUs is dependent upon the neutral appearance: for example, scoring of AU5 (upper lid raiser) is dependent upon whether the iris shows entirely in the neutral face or is partially covered.

Without face-specific calibration, automatic AU recognition systems would suffer from similar facial attribute biases for faces not seen in the training set. Therefore, we propose to perform one-frame calibration (OFC) in AU recognition: for each face, a single image of its neutral expression is used as the reference image for calibration. The most intuitive OFC method would be to directly subtract the model's estimations of AUs in the neutral image from its AU estimations of all images of the same face. However, we show that the performance improvement from this method, if any, is limited. We propose a highly effective neural network architecture for OFC, Calibrating Siamese Network (CSN), where the reference image (the neutral face) and the target image are fed into two identical networks and joined in an intermediate stage of the network by computing the difference between their feature maps of that stage. In this paper, we demonstrate its remarkable effectiveness with a simple iResNet-50 (IR50) backbone. On the DISFA \cite{mavadati2012automatic,mavadati2013disfa}, DISFA+ \cite{mavadati2016extended}, and UNBC-McMaster \cite{lucey2011painful} datasets, we show that our OFC CSN-IR50 model (a) significantly improves the performance of IR50 by mitigating facial attribute biases, (b) substantially outperforms the naive OFC method of baseline subtraction (BS) as well as (c) a fine-tuned version of this naive OFC method, 
and (d) also outperforms state-of-the-art (SOTA) NCG models.

\section{Related Work}

\subsection{AU Recognition}

Automatic AU recognition has advanced rapidly in recent years. Traditional AU recognition methods mostly relied on classification of hand-crafted features \cite{valstar2006fully,zhao2015joint,baltruvsaitis2015cross} and have limitations. With the rise of deep learning, a large number of neural network-based methods have emerged and greatly outperformed traditional methods \cite{walecki2017deep,linh2017deepcoder,li2018eac,shao2018deep,shao2019facial,XuNIPS2019,shao2021jaa,song2021uncertain,tang2021piap,luo2022learning,chang2022knowledge,zhang2023weakly,liu2023multi,wang2024multi}. Most relevant to our work, the limited number of participants in AU datasets often causes AU recognition models to overfit to person-specific features, thereby degrading their generalization performance. To address this issue, prior studies \cite{tang2021piap,baltruvsaitis2015cross,yang2019learning} have proposed a range of approaches. In contrast, this work introduces a Siamese architecture for AU recognition.

While various datasets have been proposed for AU recognition \cite{mavadati2013disfa,mavadati2016extended,zhang2014bp4d,lucey2011painful,girard2017sayette,mcduff2018fed+}, DISFA and BP4D have been most widely used for benchmarking. Although both datasets provide intensity labels, most recent works in AU recognition only focused on AU detection (indicating only whether an AU is present in each frame) \cite{shao2021jaa,song2021uncertain,tang2021piap,luo2022learning,chang2022knowledge,zhang2023weakly,liu2023multi,wang2024multi}; only a few of them reported results on AU intensity estimation (which outputs the specific intensity levels of each AU) \cite{walecki2017deep,linh2017deepcoder,shao2019facial,fan2020facial}. Therefore, although we analyzed the performance of our proposed method on both tasks, we can compare it with the performances of only a few existing methods in AU intensity estimation.

\subsection{Bias in Facial Expression Recognition}
Previous studies have shown that facial expression recognition systems can exhibit biases across groups based on gender \cite{domnich2021responsible}, race \cite{sham2023ethical}, age \cite{kim2021age}, among other factors \cite{raina22,churamani2022domain}, raising significant attention and concerns about the fairness of these systems. Due to the objectivity of FACS by definition, some researchers believe AU recognition is less subject to bias and use it to investigate and mitigate bias in facial emotion annotation \cite{chen2021understanding} and recognition \cite{suresh2022using}. However,  AU recognition is also subject to bias \cite{Fabi22,raina22,Monares23}, which researchers have developed methods to mitigate \cite{kara2021towards,churamani2022domain}.

Although group-based biases attract more attention, they are essentially specific manifestations of the broader issue of identity bias in facial expression recognition. Researchers have proposed three methods of addressing identity bias: developing identity-aware/personalized models for facial expression recognition \cite{ meng2017identity,tu2019idennet,zhang2020identity,Xu2021personalized}, conducting face normalization before expression recognition \cite{hernandez2022deepfn}, or applying adversarial training with respect to identities on the models to encourage them to disregard identity-related features \cite{zhang2018identity}.

\subsection{Siamese Neural Network}

The Siamese Neural Network was first introduced by \cite{bromley1993signature} and has been widely applied in facial identity-related tasks, such as face verification \cite{taigman2014deepface} and face recognition \cite{yang2017neural}.

\section{Methods}

\subsection{One-Frame Calibration}

While most existing AU recognition systems aim for cross-participant non-calibrated generalization (NCG) to unseen faces, it is crucial to take the face's neutral appearance into account in AU coding. Thus, we propose one-frame calibration (OFC) for AU recognition: for each face, a single image of its neutral expression is used as the reference image for calibration.

In offline benchmarking, OFC primarily applies to video datasets. The reference image selection is achieved by manually selecting one image from all frames with zero activation of annotated AUs for each face. The aim of manual selection is to ensure that in the selected reference image, (a) the unannotated AUs are also not activated or minimally activated, and (b) the face is at an appropriate angle and not partially occluded.

In real-life applications of AU recognition systems, the ideal method of selecting the reference image for OFC is to directly ask the user to pose a neutral face before usage. 

\subsection{Calibrating Siamese Network for OFC}

We propose the Calibrating Siamese Network (CSN) architecture for OFC. The input for CSN consists of the target image for AU recognition and the reference image. The two images are fed into two identical networks with shared weights and joined in an intermediate stage of the network by computing the difference between their feature maps.

This architecture design can be integrated with a variety of model designs for AU recognition. To demonstrate its effectiveness in a simple way, we use the classical iResNet-50 (IR50) as the backbone in this work and name it CSN-IR50.

\subsubsection{CSN-IR50}

\Cref{fig:CSN-IR50} illustrates the architecture of CSN-IR50. The reference image with a neutral expression and the target image for AU recognition are fed into two identical IR50 networks with shared weights; just before reaching stage 4 of the network, the difference between their feature maps is computed and then fed into the rest of the IR50 network until the AU intensities are outputted.

CSN-IR50 may be more precisely called CSN-IR50-Stage4, emphasizing stage 4 as the merge point, which is the main version of CSN-IR50 we primarily investigate in this work. Stage 4 is selected as the merge point because the feature maps in this stage capture high-level abstractions of the face features and still retain the fine-grained information. We will also compare it with other versions with different merge points.

\subsection{Baseline Models}

We compare the performance of our proposed method with those of various models. First, we directly compare our method of OFC with CSN-IR50 with the vanilla NCG with IR50 to demonstrate the effectiveness of OFC. Second, we compare our model with the naive OFC method of baseline subtraction (BS) (IR50 OFC w/ BS) to demonstrate the superiority of our model as an OFC method. (\Cref{tab:different_versions_of_CSN-IR50} shows a comparison with CSN-IR50-Output, which is a fine-tuned version of this naive method (fine-tuned with the same parameters as CSN-IR50)).
Additionally, we also compare  our model's performance with that of other state-of-the-art (SOTA) NCG models.
\subsubsection{IR50 (NCG)}

In IR50 (NCG), the IR50 is trained on individual images of the training set and directly applied to images of unseen faces during validation.

\subsubsection{IR50 (OFC w/ BS)}

In IR50 (OFC w/ BS), the IR50 is trained on individual images of the training set; however, during validation, its output on the reference image for each face is used as the baseline, which is subtracted from outputs on all images of the same face to produce final predictions.

\section{Experiments}

\begin{table*}[t]
    \caption{The performance of different methods on AU intensity estimation on the DISFA dataset. For each metric, the best results in each column are shown in bold. The rows below the dashed lines in each section include the three methods we propose for comparison; the best results among them are underlined.}
    \centering
    \begin{tabular}{|c|cccccccccccccc|}
        \hline
        \multirow{2}{*}{Metric} & \multirow{2}{*}{Method} & \multicolumn{12}{c}{AU} & \multirow{2}{*}{Average}\\
        \cline{3-14}
        & & 1 & 2 & 4 & 5 & 6 & 9 & 12 & 15 & 17 & 20 & 25 & 26 & \\ \hline
        \multirow{7}{*}{ICC(3,1)$\uparrow$} & 
        CCNN-IT \cite{walecki2017deep} & .18 & .15 & .61 & .07 & .65 & .55 & .82 & .44 & .37 & .28 & .77 & .54 & .45\\
        & 2DC \cite{linh2017deepcoder} & .70 & .55 & .69 & .05 & .59 & .57 & \textbf{.88} & .32 & .10 & .08 & .90 & .50 & .50\\
        & SCC-Heatmap \cite{fan2020facial} & .73 & .44 & .74 & .06 & .27 & .51 & .71 & .04 & .37 & .04 & .94 & \textbf{.78} & .47\\
        & iARL \cite{shao2019facial} & .13 & .36 & .68 & .22 & .56 & .36 & .86 & \textbf{.52} & .37 & .12 & \textbf{.96} & .60 & .48\\
        \cdashline{2-15}
        & IR50 (NCG) & .53 & .45 & .75 & .62 & .55 & .57 & .84 & \underline{.42} & .47 & .24 & .93 & .65 & .59\\
        & IR50 (OFC w/ BS) & .62 & .51 & .75 & .55 & .60 & .59 & .82 & .39 & .44 & .20 & .93 & .68 & .59\\
        & CSN-IR50 (OFC) & \underline{\textbf{.75}} & \underline{\textbf{.70}} & \underline{\textbf{.80}} & \underline{\textbf{.72}} & \underline{\textbf{.67}} & \underline{\textbf{.61}} & \underline{.85} & .33 & \underline{\textbf{.52}} & \underline{\textbf{.37}} & \underline{.94} & \underline{.77} & \underline{\textbf{.67}} \\
        \hline
        
        \multirow{6}{*}{MAE$\downarrow$} & 
        CCNN-IT \cite{walecki2017deep} & .87 & .63 & .86 & .26 & .73 & .57 & .55 & .38 & .57 & .45 & .81 & .64 & .61\\
        & SCC-Heatmap \cite{fan2020facial} & \textbf{.16} & \textbf{.16} & \textbf{.27} & \textbf{.03} & \textbf{.25} & \textbf{.13} & .32 & .15 & \textbf{.20} & .09 & .30 & .32 & \textbf{.20}\\
        & iARL \cite{shao2019facial} & .30 & .31 & .52 & .04 & .36 & .30 & \textbf{.31} & \textbf{.05} & .33 & \textbf{.08} & .29 & \textbf{.26} & .26\\
        \cdashline{2-15}
        & IR50 (NCG) & .37 & .39 & .44 & .11 & .35 & .21 & .34 & .20 & .39 & .21 & .32 & .42 & .31\\
        & IR50 (OFC w/ BS) & .30 & .36 & .41 & .14 & .33 & .20 & .40 & .18 & .37 & .31 & .34 & .38 & .31\\
        & CSN-IR50 (OFC) & \underline{.19} & \underline{\textbf{.16}} & \underline{.38} & \underline{.08} & \underline{.26} & \underline{.19} & \underline{\textbf{.31}} & \underline{.17} & \underline{.22} & \underline{.13} & \underline{\textbf{.27}} & \underline{.27} & \underline{.22} \\
        \hline

    \end{tabular}
    \label{tab:AU_intensity_estimation_performances_on_DISFA}
\end{table*}

\subsection{Datasets and Settings}

\textbf{DISFA} \cite{mavadati2013disfa} contains left-view and right-view facial video recordings of 27 participants with approximately 130,000 frames in total for each view. Each frame is annotated with intensities of 12 AUs on a scale of 0 to 5. Following previous studies, we perform participant-exclusive 3-fold cross-validation on DISFA.

\textbf{DISFA+} \cite{mavadati2016extended} is an extension of the DISFA dataset. It contains facial video recordings of 9 participants’ posed and spontaneous facial expressions with each frame being annotated with the same 12 AUs on a scale of 0 to 5. We perform leave-one-participant-out cross-validation on DISFA+.

\textbf{UNBC-McMaster} \cite{lucey2011painful} is a dataset originally collected for pain detection. Since it also contains frame-level AU intensity annotations of 10 AUs on a scale of 0 to 5 (except that the annotations for AU43 (eye closure) are binary), it is also appropriate for analyzing AU recognition methods. It contains facial video recordings of 25 participants with 48,398 frames in total. We perform participant-exclusive 5-fold cross-validation on UNBC-McMaster.

The three datasets collectively include the following AUs: AU1 (Inner Brow Raiser), AU2 (Outer Brow Raiser), AU4 (Brow Lowerer), AU5 (Upper Lid Raiser), AU6 (Cheek Raiser), AU7 (Lid Tightener), AU9 (Nose Wrinkler), AU10 (Upper Lip Raiser), AU12 (Lip Corner Puller), AU15 (Lip Corner Depressor), AU17 (Chin Raiser), AU20 (Lip Stretcher), AU25 (Lips Part), AU26 (Jaw Drop), and AU43 (Eye Closure).

We did not include the widely used BP4D \cite{zhang2014bp4d} dataset because it is not appropriate for OFC. Unlike the previously mentioned datasets, in BP4D, FACS coders selectively annotate a 20-second segment with the highest density of facial expression for each recording session, and only these segments are released in the dataset. Consequently, for most participants in BP4D, there is no appropriate ``neutral face frame'' to use as the reference image for OFC.

We evaluate our methods on both AU intensity estimation and AU detection. In AU intensity estimation, the model outputs estimations of intensities of the AUs (real values between 0 and 5). In AU detection, the model outputs predictions of whether each AU occurs in binary format (0 or 1). Following previous studies \cite{shao2019facial}, for AU detection, we consider AU intensities greater than or equal to 2 as occurrences, and we only include 8 of the 12 AUs for DISFA.

\subsection{Implementation Details}

Each frame is preprocessed with face detection \cite{lugaresi2019mediapipe}, face alignment \cite{PFL}, and a combination of histogram equalization and linear mapping \cite{kuo2018compact} for both training and validation.

We use the weights pre-trained on Glint360k
\cite{deng2019arcface,an2022killing} for both IR50 and CSN-IR50, and the last layer of the network is modified to adapt to the output format for the task.

For AU intensity estimation, we train the network to perform both regression and ordinal classification \cite{niu2016ordinal} on the AU intensities. The network outputs the estimation of the AUs in two formats: for estimating the intensity of the $i$th AU $y_{i}$, it outputs 1 value $\hat{y}_{i, \mathrm{reg}}$ representing the numerical estimation of the AU intensity (in the format of regression) and 5 values $\sigma(\hat{y}_{i, \mathrm{class(1)}})$, $\sigma(\hat{y}_{i, \mathrm{class(2)}})$, $\sigma(\hat{y}_{i, \mathrm{class(3)}})$, $\sigma(\hat{y}_{i, \mathrm{class(4)}})$, and $\sigma(\hat{y}_{i, \mathrm{class(5)}})$ respectively representing the estimated probability of the AU intensity being higher than or equal to 1, 2, 3, 4, and 5 (in the format of binary classifications). The loss function consists of three parts:
\begin{equation}
    E_\mathrm{AUIE} = E_\mathrm{reg,MSE} + E_\mathrm{reg,cos} + E_\mathrm{class},
\end{equation}
where $E_\mathrm{reg,MSE}$, $E_\mathrm{reg,cos}$, and $E_\mathrm{class}$ respectively represent a mean squared error (MSE) loss for the numerical estimations
\begin{equation}
    E_\mathrm{reg,MSE} = \Sigma_{i=1}^{n}w_{i,y_{i}}(y_{i}-\hat{y}_{i, \mathrm{reg}})^{2},
    \label{eqn:loss_reg_MSE}
\end{equation}
a cosine similarity loss for the numerical estimations
\begin{equation}
    E_\mathrm{reg,cos} = 1 - \frac{\Sigma_{i=1}^{n}y_{i}\hat{y}_{i, \mathrm{reg}}}{(\Sigma_{i=1}^{n}{y_{i}^{2}})(\Sigma_{i=1}^{n}\hat{y}_{i, \mathrm{reg}}^{2})},
    \label{eqn:loss_reg_cos}
\end{equation}
and a cross entropy loss for the binary classification estimations
\begin{equation}
    E_\mathrm{class} = \Sigma_{i=1}^{n}\Sigma_{j=1}^{5}w_{i,j,\chi_{y_{i}\geq j}}CE(\chi_{y_{i}\geq j}, \sigma(\hat{y}_{i, \mathrm{class(j)}})),
    \label{eqn:loss_class}
\end{equation}
with the cross entropy function being 
\begin{equation}
    CE(y,\hat{y})=-[y_i \log(\hat{y}_i) + (1 - y_i) \log(1 - \hat{y}_i)].
    \label{eqn:cross_entropy}
\end{equation}
The weights for the MSE loss and those for the cross entropy loss are both inverse-frequency weighted and normalized within each AU for addressing class imbalance in the datasets (substantially higher number of occurrences for low AU intensities). 
However, since the occurrences of high intensities are too few for most AUs 
(resulting in too high weights for the MSE loss if used directly), 
we ``bin'' the intensities into 2 groups, and each group shares the same weight. Specifically, for the MSE loss, we apply one weight for occurrences of intensities of 0 and 1 and another weight for occurrences of intensities of 2, 3, 4, and 5, and these weights are computed based on the total number of occurrences within each intensity group.

Notably, although we train the network to learn both numerical estimations and binary classification estimations of the AU intensities, only the numerical estimations are used in model validation (and any further model inference).

For AU detection, we train the network to directly output the estimated probability of the occurrence of each AU $\sigma(\hat{y}_{i})$, with an occurrence defined as $y_{i}\geq 2$. The loss function uses cross entropy loss:
\begin{equation}
    E_\mathrm{AUD} = \Sigma_{i=1}^{n}w_{i,\chi_{y_{i}\geq 2}}CE(\chi_{y_{i}\geq 2}, \sigma(\hat{y}_{i})),
    \label{eqn:loss_AU_detection}
\end{equation}
with similar inverse-frequency weights normalized within each AU. The detailed equations for weight computation in both AU intensity estimation and AU detection are provided in the technical appendix.

For model training, we employ the Adam optimizer with an initial learning rate of $10^{-4}$ for parameters of the last layer and $10^{-5}$ for other parameters, a weight decay of $5\times 10^{-4}$, and a batch size of 64. These hyperparameters were selected based on our prior work on other models for AU recognition. For each fold of each dataset, we train each model for 3 epochs with the random seed 42 in a single run using PyTorch version 2.0.0 in Python version 3.9.7, as initial exploration showed that, with the pre-trained weights, performance does not significantly change after the first epoch. For all training/validation on DISFA, we used a single NVIDIA GeForce GTX 1080Ti 11G GPU and an Intel\textsuperscript{\textregistered} Core\texttrademark{} i9-7900X CPU with 128\,GB RAM; for all training/validation on DISFA+ and UNBC-McMaster, we used a single NVIDIA RTX A6000 GPU and dual AMD EPYC 7302 16-core CPUs with 512\,GB of 8-channel RAM.

\subsection{Results of CSN-IR50}

\subsubsection{Main Results}

\begin{table*}[t]
    \caption{The performance of different methods on AU detection on the DISFA dataset. For each metric, the best results in each column are shown in bold. The rows below the dashed lines in each of the F1 score and accuracy sections include the three methods we propose for comparison; the best results among them are underlined. For precision and recall, the better results between IR50 (NCG) and CSN-IR50 (OFC) are also underlined.}
    \label{tab:AU_detection_performances_on_DISFA}
    \begin{adjustbox}{max width=\textwidth}
    \centering
    \begin{tabular}{|c|cccccccccc|}
        \hline
        \multirow{2}{*}{Metric} & \multirow{2}{*}{Method} & \multicolumn{8}{c}{AU} & \multirow{2}{*}{Average}\\
        \cline{3-10}
        & & 1 & 2 & 4 & 6 & 9 & 12 & 25 & 26 & \\
        \hline
        \multirow{11}{*}{F1 score$\uparrow$}
        & OF-Net \cite{yang2019learning} & 30.9 & 34.7 & 63.9 & 44.5 & 31.9 & 78.3 & 84.7 & 60.5 & 53.7 \\
        & ARL \cite{shao2019facial} & 43.9 & 42.1 & 63.6 & 41.8 & 40.0 & 76.2 & 95.2 & 66.8 & 58.7\\
        & UGN-B \cite{song2021uncertain} & 43.3 & 48.1 & 63.4 & 49.5 & 48.2 & 72.9 & 90.8 & 59.0 & 60.0\\
        & J\^AA-Net \cite{shao2021jaa} & 62.4 & 60.7 & 67.1 & 41.1 & 45.1 & 73.5 & 90.9 & \textbf{67.4} & 63.5\\
        & PIAP \cite{tang2021piap} & 50.2 & 51.8 & 71.9 & 50.6 & 54.5 &  \textbf{79.7} & 94.1 & 57.2 & 63.8\\
        & ME-GraphAU \cite{luo2022learning} & 54.6 & 47.1 & 72.9 & \textbf{54.0} & \textbf{55.7} & 76.7 & 91.1 & 53.0 & 63.1\\
        & KDSRL \cite{chang2022knowledge} & 60.4 & 59.2 & 67.5 & 52.7 & 51.5 & 76.1 & 91.3 & 57.7 & 64.5\\
        & CLEF \cite{zhang2023weakly} & 64.3 & 61.8 & 68.4 & 49.0 & 55.2 & 72.9 & 89.9 & 57.0 & 64.8\\
        & SACL \cite{liu2023multi} & 62.0 & \textbf{65.7} & 74.5 & 53.2 & 43.1 & 76.9 & \textbf{95.6} & 53.1 & 65.5\\
        & MDHR \cite{wang2024multi} & \textbf{65.4} & 60.2 & \textbf{75.2} & 50.2 & 52.4 & 74.3 & 93.7 & 58.2 & 66.2\\
        \cdashline{2-11}
        & IR50 (NCG) & 35.4 & 33.0 & 64.4 & 48.6 & \underline{51.8} & 77.1 & 91.9 & 58.1 & 57.5\\
        & IR50 (OFC w/ BS) & 16.3 & 17.4 & 36.1 & 21.1 & 11.5 & 39.1 & 61.8 & 26.6 & 28.7\\
        & CSN-IR50 (OFC) & \underline{65.3} & \underline{58.3} & \underline{70.8} & \underline{52.6} & 51.7 & \underline{77.3} & \underline{94.6} & \underline{65.4} & \underline{\textbf{67.0}} \\
        \hline
        \multirow{7}{*}{Accuracy$\uparrow$}
        & OF-Net \cite{yang2019learning} & 84.7 & 90.6 & 72.1 & 72.8 & 87.9 & 82.6 & 78.8 & 71.4 & 80.1 \\
        & ARL \cite{shao2019facial} & 92.1 & 92.7 & 88.5 & 91.6 & 95.9 & \textbf{93.9} & 97.3 & 94.3 & 93.3\\
        & UGN-B \cite{song2021uncertain} & 95.1 & 93.2 & 88.5 & \textbf{93.2} & \textbf{96.8} & 93.4 & 94.8 & 93.8 & 93.4\\
        & J\^AA-Net \cite{shao2021jaa} & \textbf{97.0} & \textbf{97.3} & 88.0 & 92.1 & 95.6 & 92.3 & 94.9 & \textbf{94.8} & 94.0\\
        & SACL \cite{liu2023multi} & 96.1 & 96.9 & \textbf{92.5} & 91.7 & 95.0 & 93.7 & \textbf{97.5} & 89.1 & 94.1\\
        \cdashline{2-11}
        & IR50 (NCG) & 87.1 & 87.1 & 86.8 & 89.5 & \underline{95.5} & 93.2 & 95.3 & 89.6 & 90.5\\
        & IR50 (OFC w/ BS) & 50.5 & 62.5 & 46.3 & 42.3 & 37.2 & 60.1 & 65.8 & 55.5 & 52.5\\
        & CSN-IR50 (OFC) & \underline{96.9} & \underline{96.9} & \underline{90.4} & \underline{91.7} & 94.6 & \underline{93.5} & \underline{96.9} & \underline{92.8} & \underline{\textbf{94.2}} \\
        \hline
        \multirow{2}{*}{Precision$\uparrow$ / Recall$\uparrow$}
        & IR50 (NCG)
        & 23.6 / \underline{71.0}
        & 21.2 / \underline{73.9}
        & 54.8 / \underline{78.0}
        & 39.7 / \underline{62.6}
        & \underline{46.8} / 58.1
        & 67.9 / \underline{89.2}
        & 89.0 / 94.9
        & 45.0 / \underline{81.8}
        & 48.5 / \underline{76.2} \\
        & CSN-IR50 (OFC)
        & \underline{73.1} / 59.0
        & \underline{69.1} / 50.4
        & \underline{65.7} / 76.8
        & \underline{47.7} / 58.7
        & 41.2 / \underline{69.5}
        & \underline{70.1} / 86.3
        & \underline{92.5} / \underline{96.8}
        & \underline{56.9} / 76.8
        & \underline{64.5} / 71.8 \\
        \hline

    \end{tabular}
    \end{adjustbox}
    
\end{table*}

\Cref{tab:AU_intensity_estimation_performances_on_DISFA} reports the performance of CSN-IR50 on AU intensity estimation on DISFA in comparison to other methods. Firstly, CSN substantially improves the performance of IR50 for NCG and also greatly outperforms the naive OFC method of IR50 with BS, with a difference of 0.08 in ICC(3,1) and a difference of 0.09 in Mean Absolute Error (MAE). Secondly, in comparison to other SOTA NCG methods, our CSN-IR50 demonstrates a substantially higher ICC(3,1) of 0.67 and a near-best MAE of 0.22 (the best being 0.20). ICC(3,1) measures the consistency between the model estimations and the human expert labels in the dataset. 
We believe  ICC(3,1) is a better metric here because of the high imbalance of DISFA.

\Cref{tab:AU_detection_performances_on_DISFA} reports the performance of CSN-IR50 on AU detection on DISFA in comparison to other models. Firstly, it again substantially improves the performance of IR50 for NCG and outperforms the naive OFC method of IR50 with BS\footnote{IR50 (OFC w/ BS) has substantially worse performance than other methods because baseline subtraction is intrinsically not appropriate for outputting AU occurrences in binary format. The IR50 network output for a neutral face is supposed to be very close to that for the reference image, either slightly higher or slightly lower. Thus, after baseline subtraction, the final output might be either a small positive or a small negative. When it is a small positive, it would be considered as high consistency in AU intensity estimation but would be considered as a false positive in AU detection.}
in both F1 score and accuracy. Secondly, in comparison to other SOTA NCG methods, our CSN-IR50 demonstrates both a higher F1 score and a higher accuracy.

Note that the comparison between our CSN-IR50 and other SOTA models here is not an apples-to-apples comparison because CSN-IR50 is for OFC while the SOTA models are for NCG. However, also note that the effectiveness of our proposed CSN architecture is demonstrated only using the simple IR50 as the backbone for simplicity in our paper, and we believe the CSN architecture design has great potential to be integrated with more complicated, advanced backbone models for AU recognition to achieve higher performance.

\begin{table}[tb]
    \caption{Performance of different methods 
    on the DISFA+ and UNBC-McMaster datasets. The results shown here are all average values across all AUs. The best results in each column are underlined.}
    \centering
    \tablesize
    \begin{tabular}{|c|cc|cc|}
        \hline
        \multirow{2}{*}{Method} & \multicolumn{2}{c|}{AU Intensity Estimation} & \multicolumn{2}{c|}{AU Detection} \\
        \cline{2-5}
         & ICC(3,1)$\uparrow$ & MAE$\downarrow$ & F1$\uparrow$ & Accuracy$\uparrow$\\
        \hline
        \multicolumn{5}{|c|}{DISFA+}\\
        \hline
        IR50 (NCG) & .81 & .37 & 67.3 & 91.7\\
        IR50 (OFC w/ BS) & .83 & .32 & 28.3 & 47.4\\
        CSN-IR50 (OFC) & \underline{.86} & \underline{.23} & \underline{78.6} & \underline{96.2}\\
        \hline
        \multicolumn{5}{|c|}{UNBC-McMaster}\\
        \hline
        IR50 (NCG) & .30 & .29 & 25.9 & 93.3 \\
        IR50 (OFC w/ BS) & .34 & .23 & 9.2 & 36.1\\
        CSN-IR50 (OFC) &  \underline{.45} & \underline{.20} & \underline{34.2} & \underline{95.9} \\
        \hline
    \end{tabular}
    \label{tab:AU_recognition_performances_on_DISFA+_and_UNBC-McMaster}
\end{table}

As shown in \Cref{tab:AU_recognition_performances_on_DISFA+_and_UNBC-McMaster}, our CSN-IR50 demonstrates similar superiority over IR50 (NCG) and IR50 (OFC w/ BS) on the DISFA+ and UNBC-McMaster datasets. No results of other methods are shown because both datasets have rarely been used for evaluating recent SOTA AU recognition models.

\subsubsection{The Advantage of OFC with CSN}
\begin{table}[tb]
    \caption{Comparison of within-participant ICC(3,1), averaged across
    participants, and across-participant ICC(3,1) for AU intensity
    estimation. Across-participant ICC(3,1) is used elsewhere because it
    more effectively captures bias across participants. All values are
    averaged across AUs. The best result in each column
    setting is underlined.}
    \label{tab:across_vs_within_participant_ICC}
    \centering
    \tablesize
    \begin{tabular}{|c|cc|}
        \hline
        \multirow{2}{*}{Method} & \multirow{2}{*}{\begin{tabular}{c}
             Across-Participant \\ ICC(3,1)$\uparrow$
        \end{tabular}} & \multirow{2}{*}{\begin{tabular}{c}
             Within-Participant \\ ICC(3,1)$\uparrow$
        \end{tabular}}\\ & & \\
        \hline
        \multicolumn{3}{|c|}{DISFA}\\
        \hline
        IR50 (NCG) & .59 & .51\\
        CSN-IR50 (OFC) & \underline{.67} & \underline{.53}\\
        \hline
        \multicolumn{3}{|c|}{DISFA+}\\
        \hline
        IR50 (NCG) & .81 & .84\\
        CSN-IR50 (OFC) & \underline{.86} & \underline{.85}\\
        \hline
        \multicolumn{3}{|c|}{UNBC-McMaster}\\
        \hline
        IR50 (NCG) & .30 & .22\\
        CSN-IR50 (OFC) & \underline{.45} & \underline{.26}\\
        \hline
    \end{tabular}
\end{table}

\begin{figure*}[htb!]
    \centering
    \begin{subfigure}{\linewidth}
    \centering
    \includegraphics[width=.9\linewidth]
    {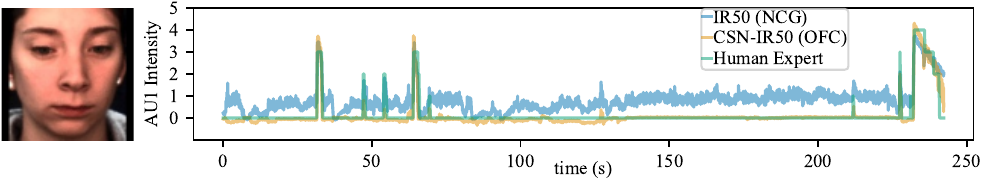}
        \caption{CSN-IR50 (OFC) mitigates bias in AU1 (inner brow raiser) intensity estimation due to wider eyebrow-to-eye distances.}
        \label{fig:example_bias_mitigation_AU1}
    \end{subfigure}
    \begin{subfigure}{\linewidth}
        \centering
        \includegraphics[width=.9\linewidth]{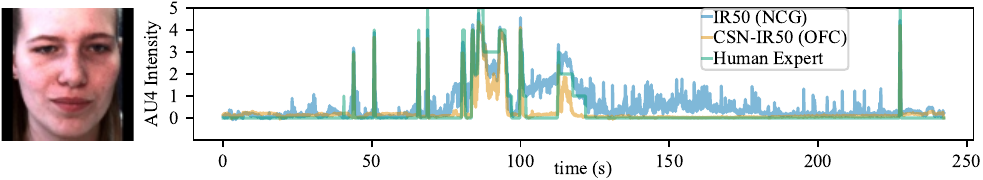}
        \caption{CSN-IR50 (OFC) mitigates bias in AU4 (brow lowerer) intensity estimation due to the wrinkle.}
         \label{fig:example_bias_mitigation_AU4}
    \end{subfigure}
    \begin{subfigure}{\linewidth}
        \centering
        \includegraphics[width=.9\linewidth]{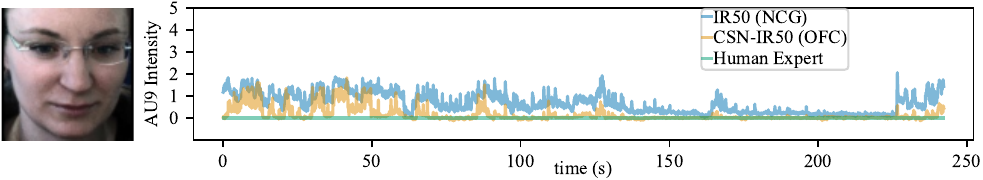}
        \caption{CSN-IR50 (OFC) mitigates bias in AU9 (nose wrinkler) intensity estimation due to the bridge of eyeglasses.}
        \label{fig:example_bias_mitigation_AU9}
    \end{subfigure}
    \begin{subfigure}{\linewidth}
        \centering\includegraphics[width=.9\linewidth]{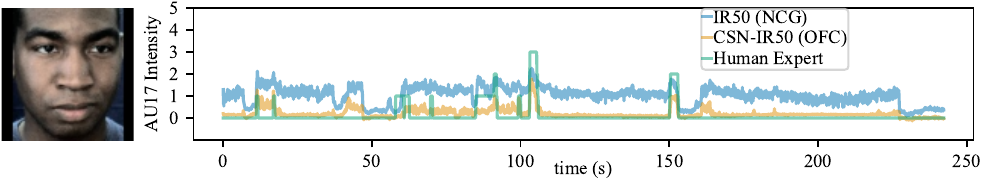}
        \caption{CSN-IR50 (OFC) mitigates bias in AU17 (chin raiser) intensity estimation due to the facial hair.}
        \label{fig:example_bias_mitigation_AU17}
    \end{subfigure}
    \caption{Examples of CSN-IR50 mitigating facial attribute biases in AU intensity estimation from the DISFA dataset. In each subfigure, the left panel is the (preprocessed) reference image of the specific participant, and the right panel is the comparison between AU intensities estimated by IR50 (NCG) and CSN-IR50 (OFC) and the human-expert-labeled intensities in the dataset.}
    \label{fig:example_bias_mitigation}
\end{figure*}

One interesting question is how/why OFC with CSN-IR50 outperforms the vanilla IR50.
Two important observations provide insights. 

Firstly, \Cref{tab:across_vs_within_participant_ICC} compares within-participant ICC(3,1) averaged over all participants and across-participant ICC(3,1) between different methods on AU intensity estimation on DISFA, DISFA+, and UNBC-McMaster. Within-participant ICC(3,1) only measures the consistency between the model estimations and human expert labels within individual participants; across-participant ICC(3,1) is the version used everywhere else in this paper, as it measures the consistency not only within but also across participants and thus more insightfully captures bias across different participants. As shown in \Cref{tab:across_vs_within_participant_ICC}, although our CSN-IR50 greatly improves the across-participant ICC(3,1) of IR50, its improvement in within-participant ICC(3,1) is much more modest. This suggests that the primary advantage of OFC with CSN-IR50 lies in its ability to calibrate for diverse facial attributes across different identities, which aligns with our original intent for CSN.

The comparison of precision and recall in \Cref{tab:AU_detection_performances_on_DISFA} offers further insights into how the calibration is achieved: the CSN architecture generally increases precision while decreasing recall for most AUs. In other words, the CSN architecture reduces the misidentification of non-activated AUs as activated (false positives), although this improvement comes at the cost of missing some actual AU activations (false negatives).

This reduction of false positives is achieved through mitigating facial attribute biases. More specifically, without face-specific calibration, some facial attributes are easily misidentified as AU activations, and our CSN addresses this issue. See \Cref{fig:example_bias_mitigation} for a variety of case examples. In \Cref{fig:example_bias_mitigation_AU1}, IR50 tends to overestimate AU1 (inner brow raiser) intensities due to the participant's wider eyebrow-to-eye distances, because AU1 produces wider distances between eyebrows and eyes; in \Cref{fig:example_bias_mitigation_AU4}, IR50 tends to overestimate AU4 (brow lowerer) intensities due to the participant's slight permanent wrinkle at the root of the nose, because AU4 may produce horizontal wrinkles at the root of the nose; in \Cref{fig:example_bias_mitigation_AU9}, IR50 tends to misidentify the bridge of eyeglasses as wrinkles caused by AU9 (nose wrinkler) activation, because AU9 produces wrinkles at the root of the nose; in \Cref{fig:example_bias_mitigation_AU17}, IR50 tends to misidentify the facial hair as wrinkles caused by AU17 (chin raiser) activation, because AU17 produces wrinkles on the chin boss.
Interestingly, the first two examples are issues human FACS coders may also face without a neutral reference, while the latter two examples are facial attribute misidentification problems specific to machine learning models, partially due to insufficient training data, a limitation common in AU datasets.
CSN-IR50 addresses these issues
by calibrating the AU estimations of different faces based on their neutral appearances.

\subsubsection{Comparing Different Versions of CSN-IR50}

\begin{table}[tb]
    \caption{Performance of different versions of CSN-IR50 on AU intensity estimation and detection on the DISFA dataset. The suffix indicates where the two networks in CSN-IR50 merge; for example, CSN-IR50-Stage4 means that the two networks merge (by computing their difference) just before stage 4 of IR50. The boxed CSN-IR50-Stage4 is the main version we use in the rest of the paper (and shown in Figure \ref{fig:CSN-IR50}). The results shown here are all average values across all AUs. The best results in each column are underlined.}
    \centering
    \tablesize
    \begin{tabular}{|c|cc|cc|}
        \hline
        \multirow{2}{*}{Method} & \multicolumn{2}{c|}{AU Intensity Estimation} & \multicolumn{2}{c|}{AU Detection} \\
        \cline{2-5}
         & ICC(3,1)$\uparrow$ & MAE$\downarrow$ & F1$\uparrow$ & Accuracy$\uparrow$\\
        \hline
        CSN-IR50-Stage1 & .61 & .31 & 60.8 & 92.0 \\
        CSN-IR50-Stage2 & .63 & .29 & 65.2 & 93.3\\
        CSN-IR50-Stage3 & .66 & .26 & 62.4 & 92.5\\
        \smallfbox{CSN-IR50-Stage4} & \underline{.67} & \underline{.22} & \underline{67.0} & \underline{94.2} \\
        CSN-IR50-FC & .55 & .28 & 30.9 & 60.3\\
        CSN-IR50-Output & .54 & .28 & 29.9 & 58.5\\
        \hline
    \end{tabular}
    \label{tab:different_versions_of_CSN-IR50}
\end{table}

The CSN-IR50 we have presented so far is our main version, CSN-IR50-Stage4, in which the two networks for the reference image and the target image respectively merge just before stage 4 of IR50 (see \Cref{fig:CSN-IR50}). \Cref{tab:different_versions_of_CSN-IR50} compares it with other versions of CSN-IR50 with different merge points on DISFA. (Each version is named after the first module after the merge point.) We can see that our selected merge point, stage 4, is the optimal one providing the best performance. Merging at earlier stages (CSN-IR50-Stage1, CSN-IR50-Stage2, and CSN-IR50-Stage3) provides somewhat suboptimal performance 
but still outperforms the IR50 baselines. We believe these versions suffer from insufficient processing of individual faces before the merge but still benefit from calibration with the neutral reference. On the other hand, merging later just before the fully connected layer (CSN-IR50-FC) or directly computing the difference between the output AU estimations of the two networks (CSN-IR50-Output) provides substantially worse performance possibly because the more fine-grained information is already lost at that stage.
Note that CSN-IR50-Output is the fine-tuned version of our naive baseline OFC method (IR50 (OFC w/ BS)). Fine-tuning seems to have a small effect on this method, slightly improving MAE but reducing ICC(3,1) on AU intensity estimation and slightly improving F1 and accuracy on AU detection on DISFA. Its performance is much worse than our CSN-IR50-Stage4 model, which only differs in where the merging (difference computation) takes place.

\section{Discussion and Conclusion}

In this paper, we propose to perform OFC in AU recognition to better generalize the model to unseen faces and propose a CSN architecture design for OFC. For simplicity, we demonstrate the effectiveness of the CSN architecture with an IR50 backbone. On the DISFA, DISFA+, and UNBC-McMaster datasets, we show that our OFC CSN-IR50 model (a) substantially outperforms IR50 with NCG, (b) substantially outperforms IR50 with  the naive OFC method of BS, as well as (c) the fine-tuned version of this method we call CSN-IR50-Output (note that it only differs from our model in where the merging takes place), and (d) also outperforms SOTA NCG models for both AU intensity estimation and AU detection. We further show that the superiority of OFC with CSN-IR50 lies in its ability to calibrate for diverse facial attributes across different identities. Specifically, it substantially reduces false positives in AU recognition, albeit at the cost of increasing false negatives. With case examples, we demonstrate how the reduction of false positives is achieved through mitigating overestimation and misidentification of AUs due to various facial attribute biases, including eyebrow locations, wrinkles, eyeglasses, and facial hair.

One important note is that, while the comparison with CSN-IR50-Output and IR50 (OFC w/ BS) is fair and shows large performance improvement, the comparison between our CSN-IR50 and other SOTA models is not an apples-to-apples comparison because CSN-IR50 is for OFC, enhanced with one labeled frame of neutral expression from each participant in the validation set, while the SOTA models are for NCG. However, also note that the effectiveness of our proposed CSN architecture is demonstrated only using the simple IR50 as the backbone for simplicity in our paper. Additionally, CSN is not a replacement for existing NCG AU recognition models; rather, it is an augmentation that can be integrated with any existing model. Therefore, as an important future direction, we believe that our CSN architecture design has great potential to be integrated with more complicated, advanced backbone models for AU recognition to achieve higher performance.

Admittedly, OFC has limitations in real-world applications because it relies on obtaining a high-quality reference image, namely, a neutral face captured at an appropriate angle and free of partial occlusion. An ideal way to obtain such a reference image is to combine instructed posing with automated quality assessment. Specifically, the user can first be instructed to present a neutral facial expression with an appropriate, preferably frontal, head pose and without occlusion. An automated system can then verify whether the captured frame is a suitable neutral reference, for example by using the NCG base AU recognition model together with head pose estimation and occlusion detection modules. 
When a good neutral reference cannot be obtained, either because the application does not permit such user instruction or because the user consistently fails to provide a suitable reference after multiple attempts, one can still fall back to the NCG base AU recognition model or adopt an alternative calibration strategy, such as using a selected percentile of the user’s estimated AU intensities.

In conclusion, we propose to perform OFC with a novel CSN architecture design for AU recognition and demonstrate its remarkable effectiveness with a simple IR50 backbone. We further show that this significant improvement originates from the architecture’s ability to mitigate facial attribute biases. We also believe it has great potential to be integrated with better backbone models to achieve higher performance.

\section{Acknowledgements}

We thank Xiaojing Xu and Yuan Tang for helpful prior work. We are grateful for support from NSF IIS 1817226 and IIS 2208362 and seed funding from UC San Diego Social Sciences and the Sanford Institute for Empathy and Compassion as well as funds from the Hal\i c\i o\u{g}lu Data Science Institute Endowed Chair I, and hardware funding from NVIDIA, Adobe, and Sony.

%% file: sec/appendix.tex
\setcounter{section}{0}
\setcounter{table}{0}
\setcounter{figure}{0}
\setcounter{footnote}{0}

\renewcommand{\thesection}{\Alph{section}}
\renewcommand{\thetable}{S\arabic{table}}
\renewcommand{\thefigure}{S\arabic{figure}}

\clearpage
\setcounter{page}{1}
\maketitlesupplementary

\begin{figure*}[!htb]
  \centering
  \begin{subfigure}{0.4\linewidth}
    \centering
    \includegraphics[width=\linewidth]{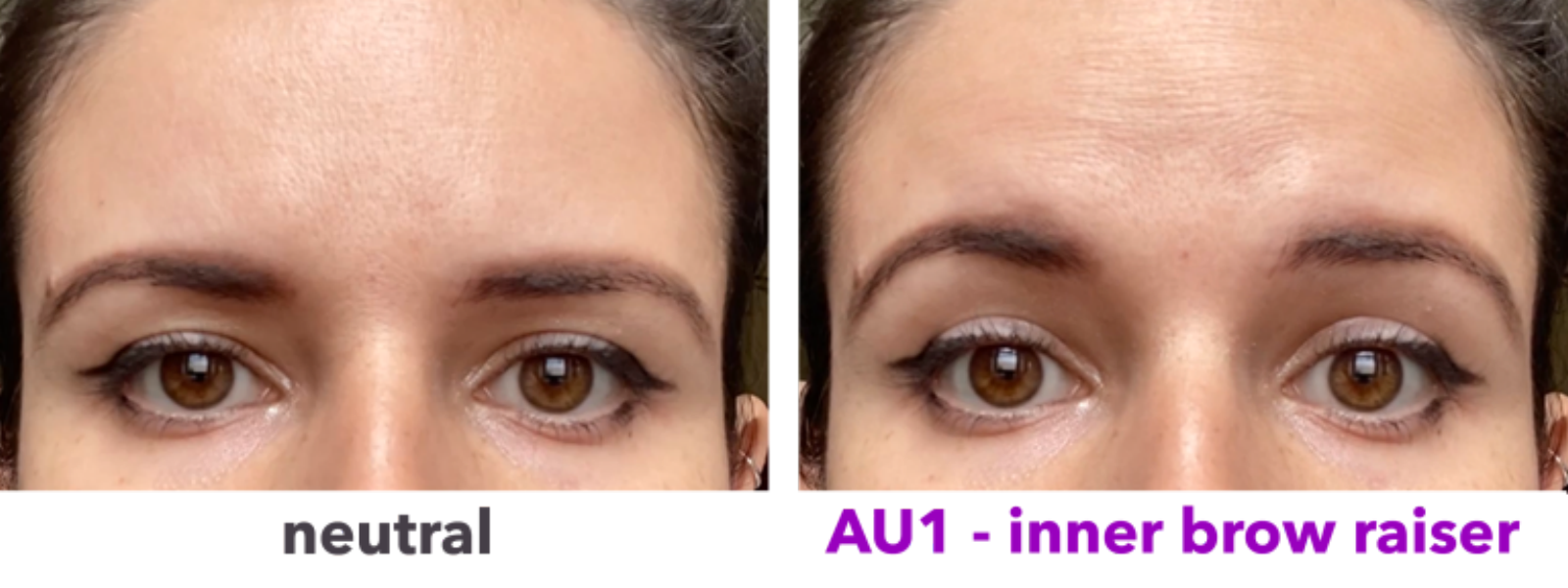}
  \end{subfigure}
  \hspace{0.08\linewidth}
  \begin{subfigure}{0.4\linewidth}
    \centering
    \includegraphics[width=\linewidth]{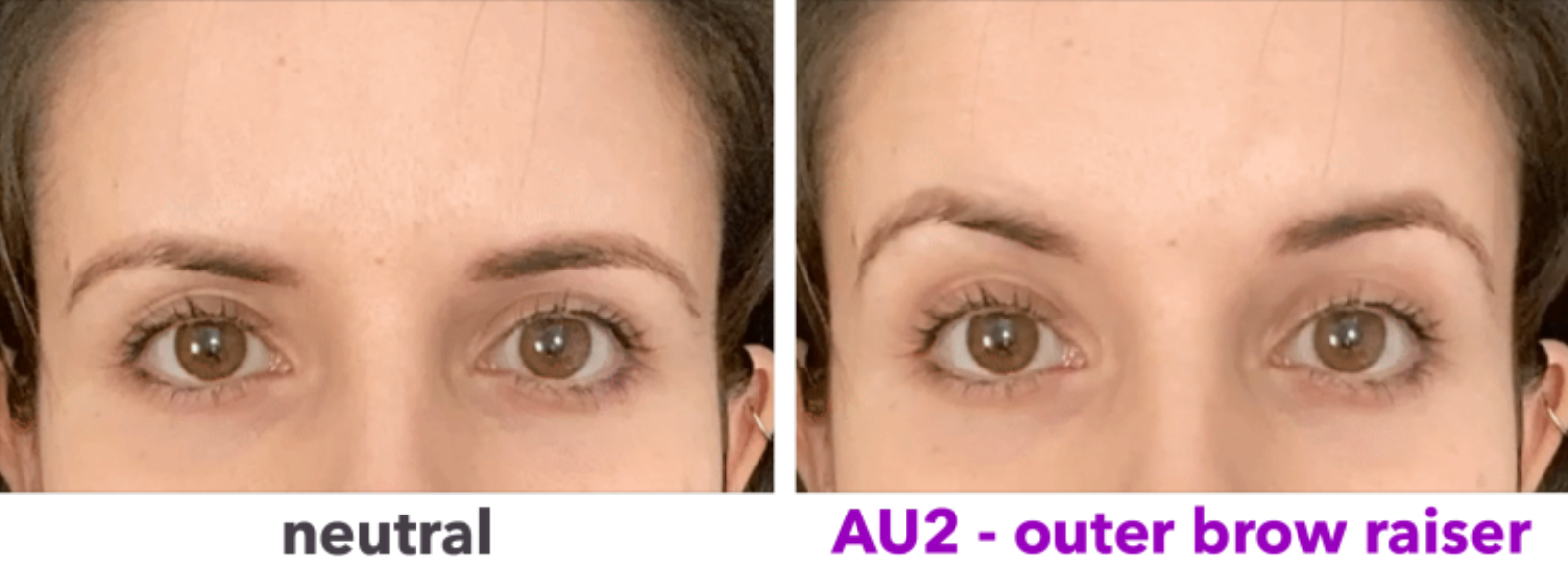}
  \end{subfigure}\\
  \begin{subfigure}{0.4\linewidth}
    \centering
    \includegraphics[width=\linewidth]{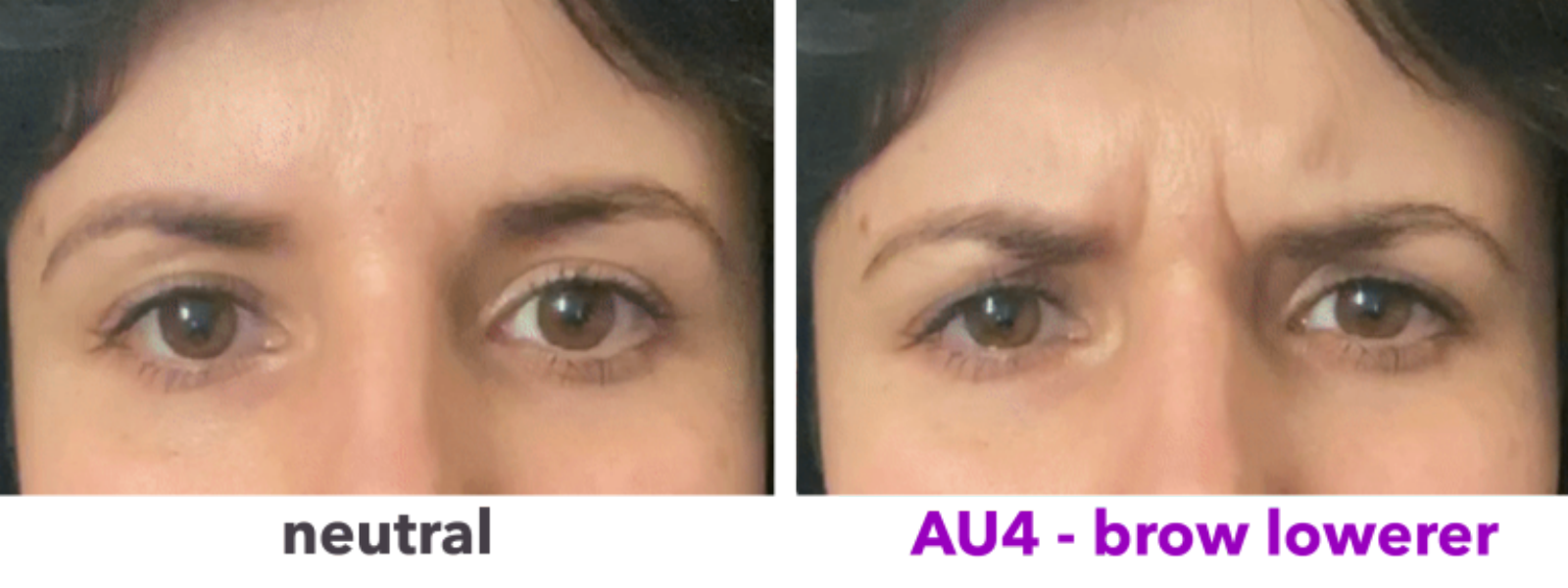}
  \end{subfigure}
  \hspace{0.08\linewidth}
  \begin{subfigure}{0.4\linewidth}
    \centering
    \includegraphics[width=\linewidth]{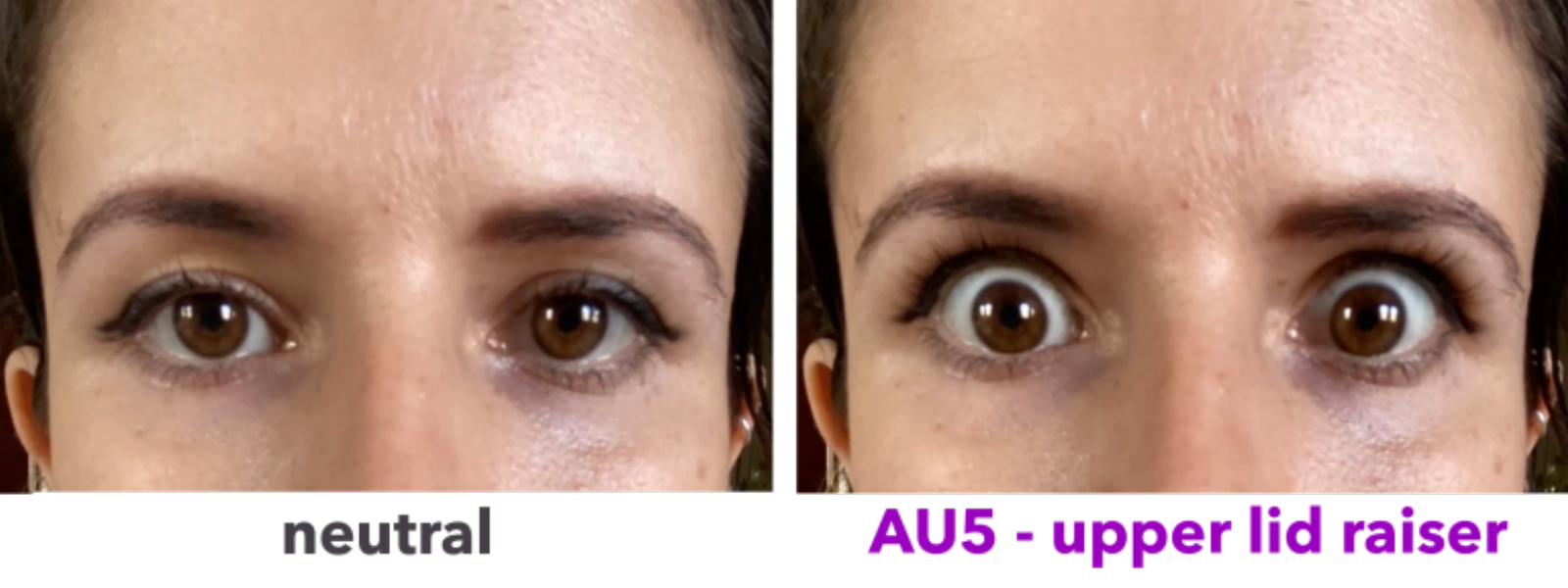}
  \end{subfigure}\\
  \begin{subfigure}{0.4\linewidth}
    \centering
    \includegraphics[width=\linewidth]{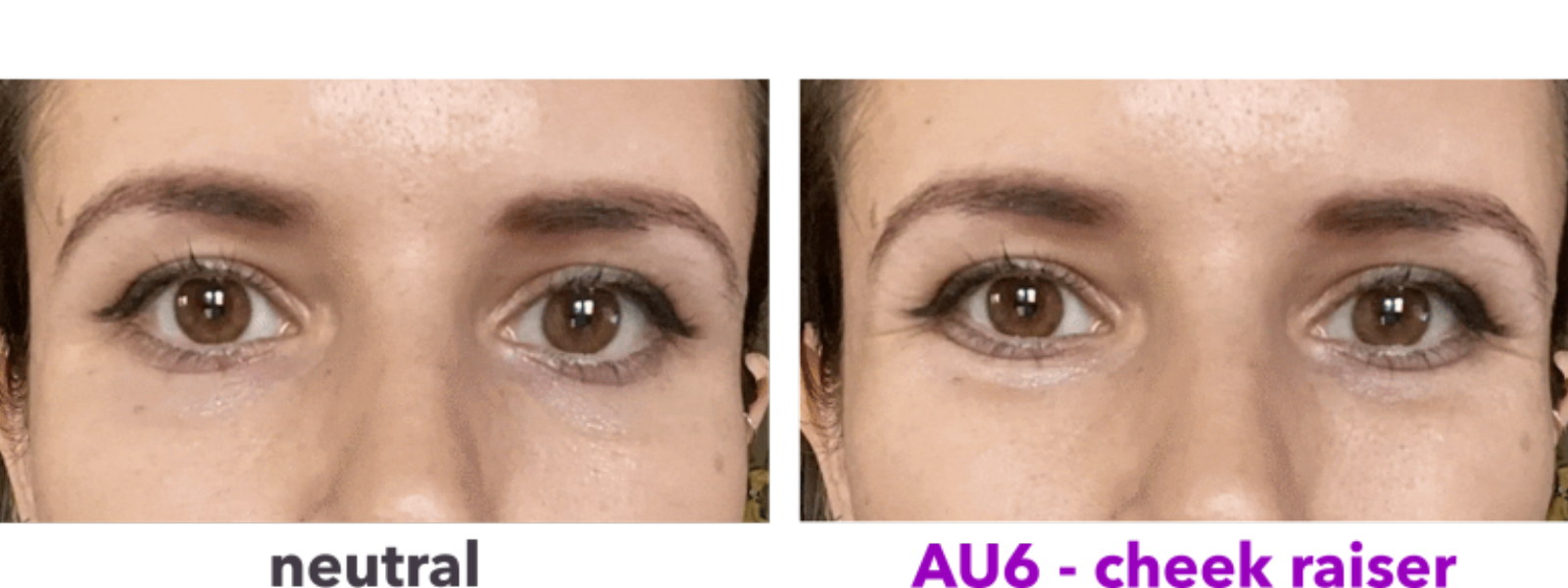}
  \end{subfigure}
  \hspace{0.08\linewidth}
  \begin{subfigure}{0.4\linewidth}
    \centering
    \includegraphics[width=\linewidth]{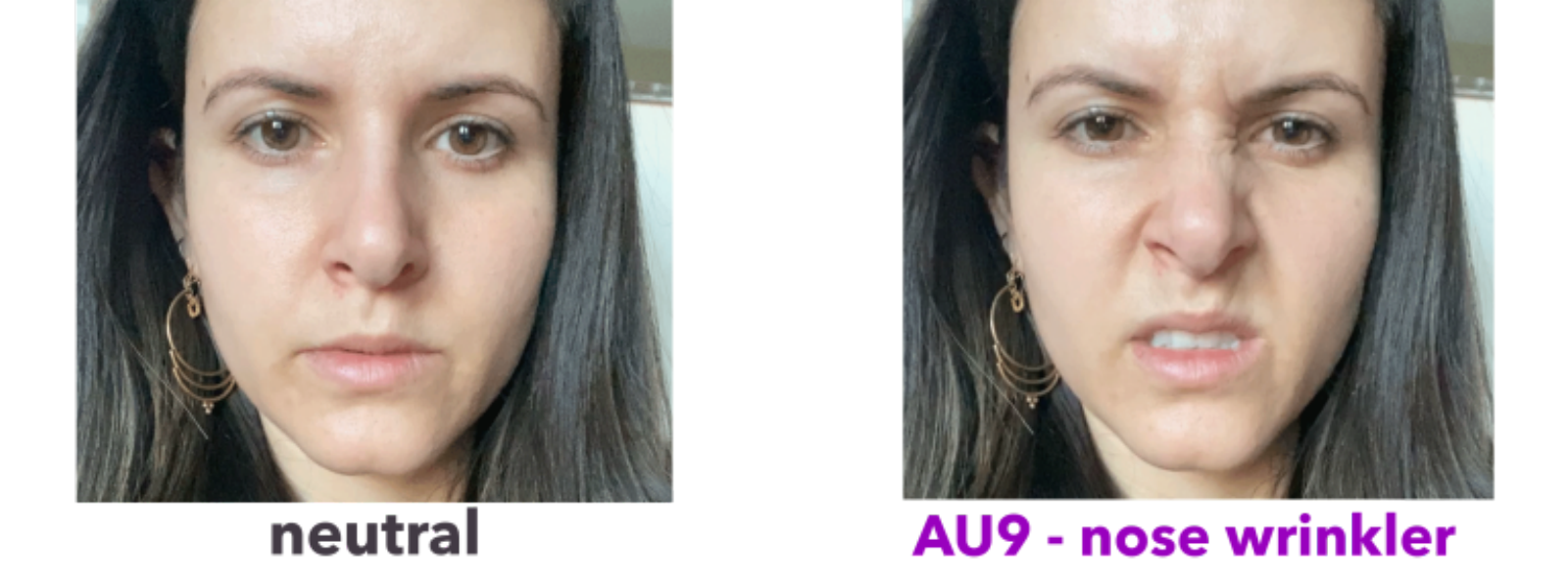}
  \end{subfigure}\\
  \begin{subfigure}{0.4\linewidth}
    \centering
    \includegraphics[width=\linewidth]{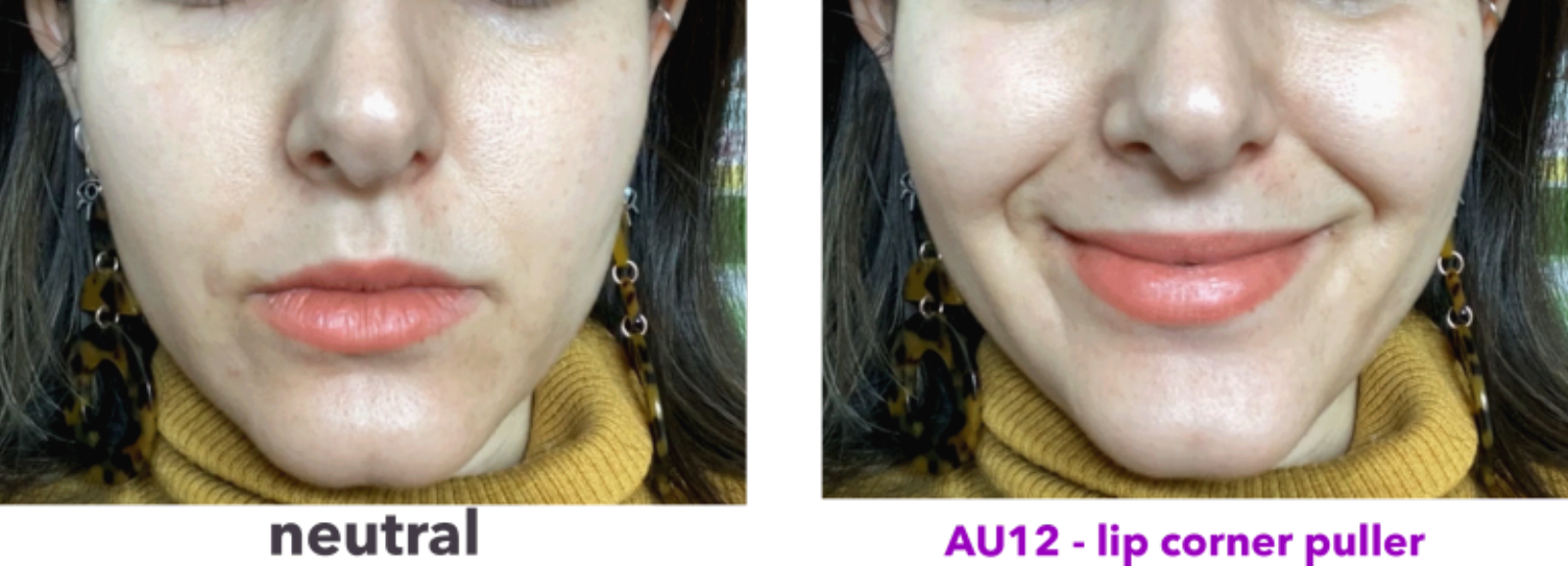}
  \end{subfigure}
  \hspace{0.08\linewidth}
  \begin{subfigure}{0.4\linewidth}
    \centering
    \includegraphics[width=\linewidth]{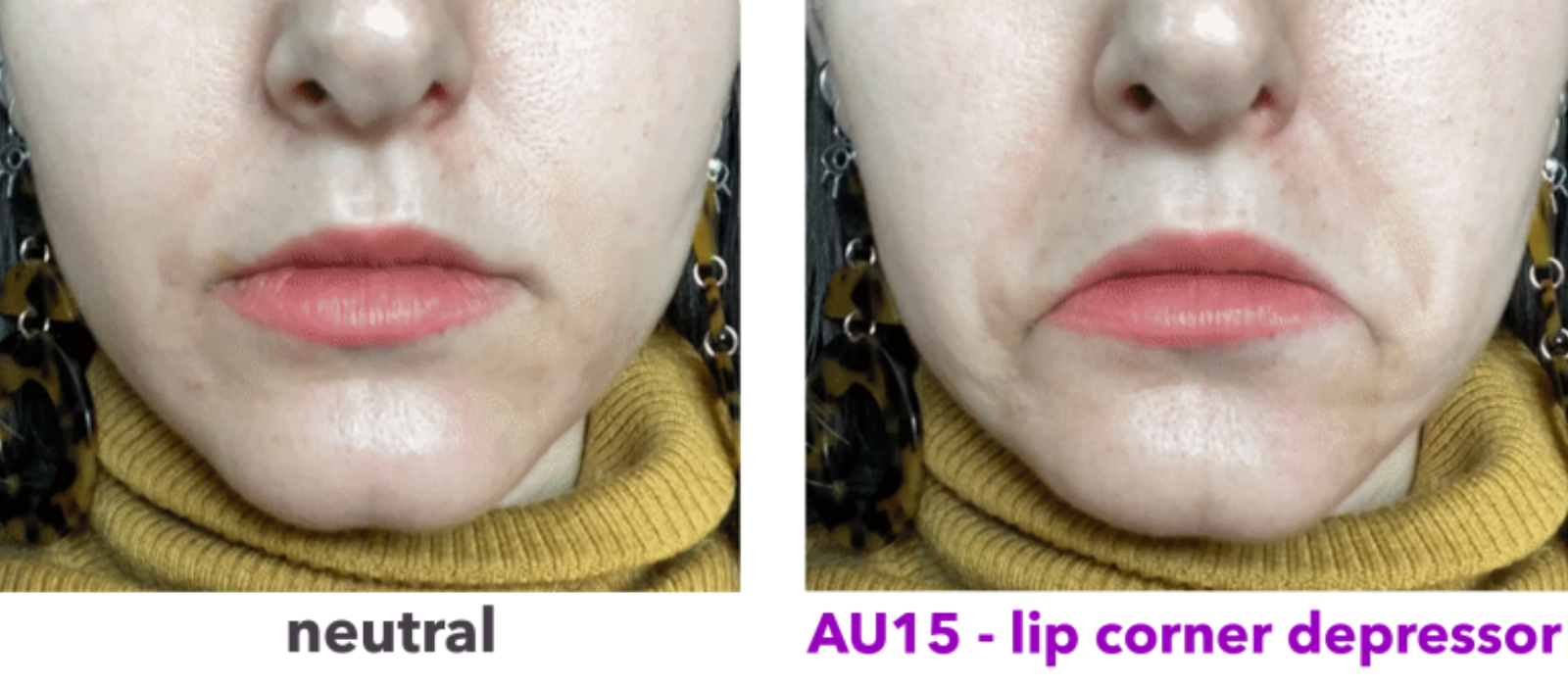}
  \end{subfigure}\\
  \begin{subfigure}{0.4\linewidth}
    \centering
    \includegraphics[width=\linewidth]{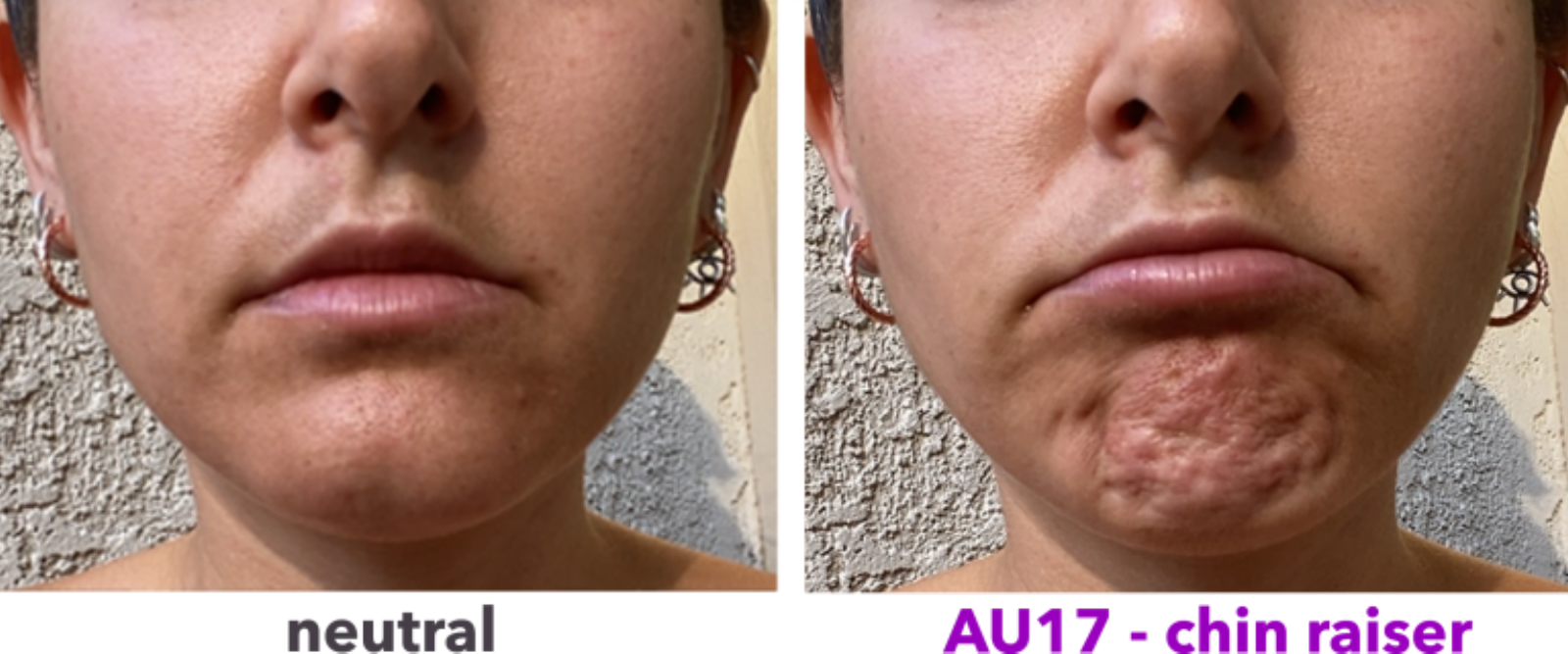}
  \end{subfigure}
  \hspace{0.08\linewidth}
  \begin{subfigure}{0.4\linewidth}
    \centering
    \includegraphics[width=\linewidth]{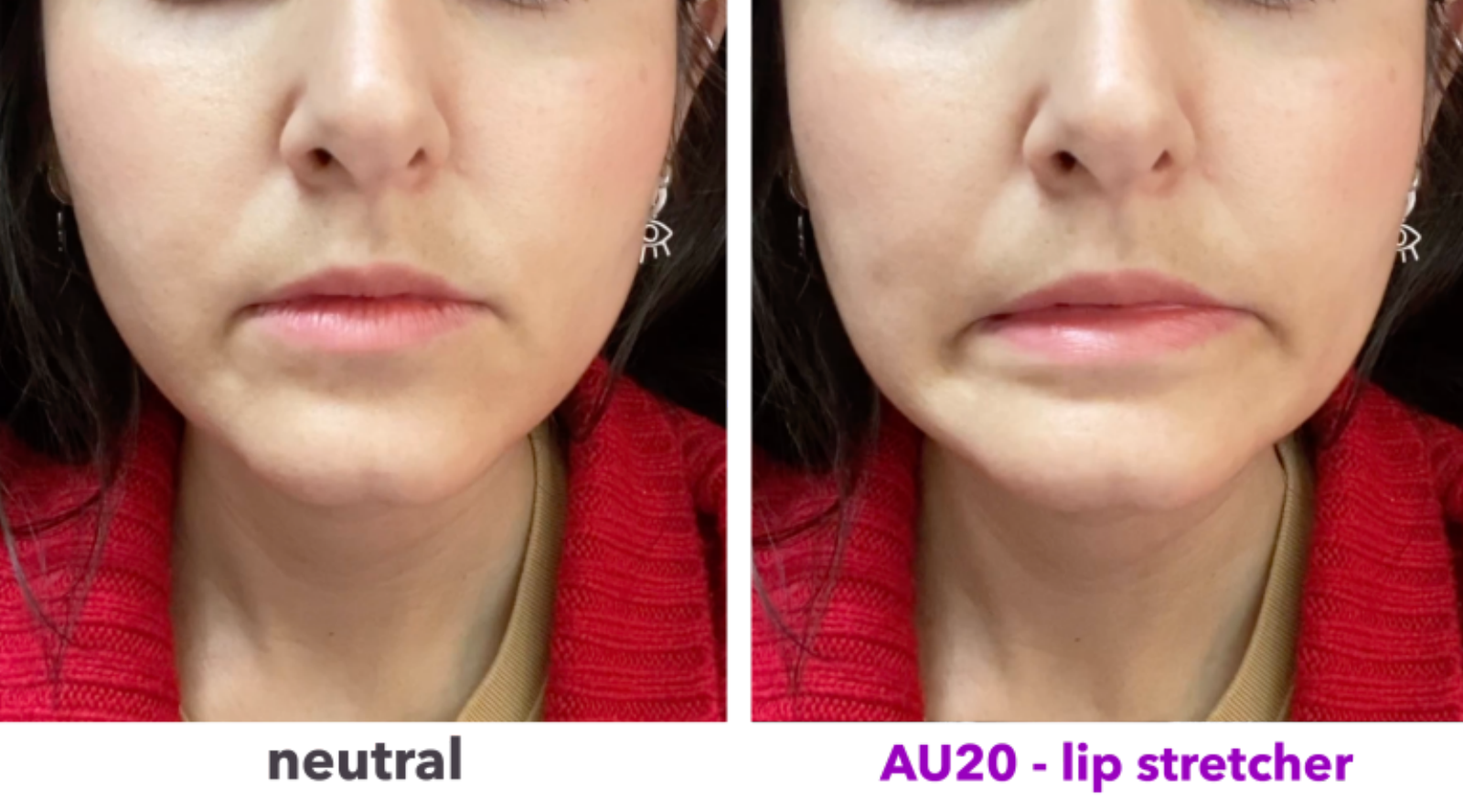}
  \end{subfigure}\\
  \begin{subfigure}{0.4\linewidth}
    \centering
    \includegraphics[width=\linewidth]{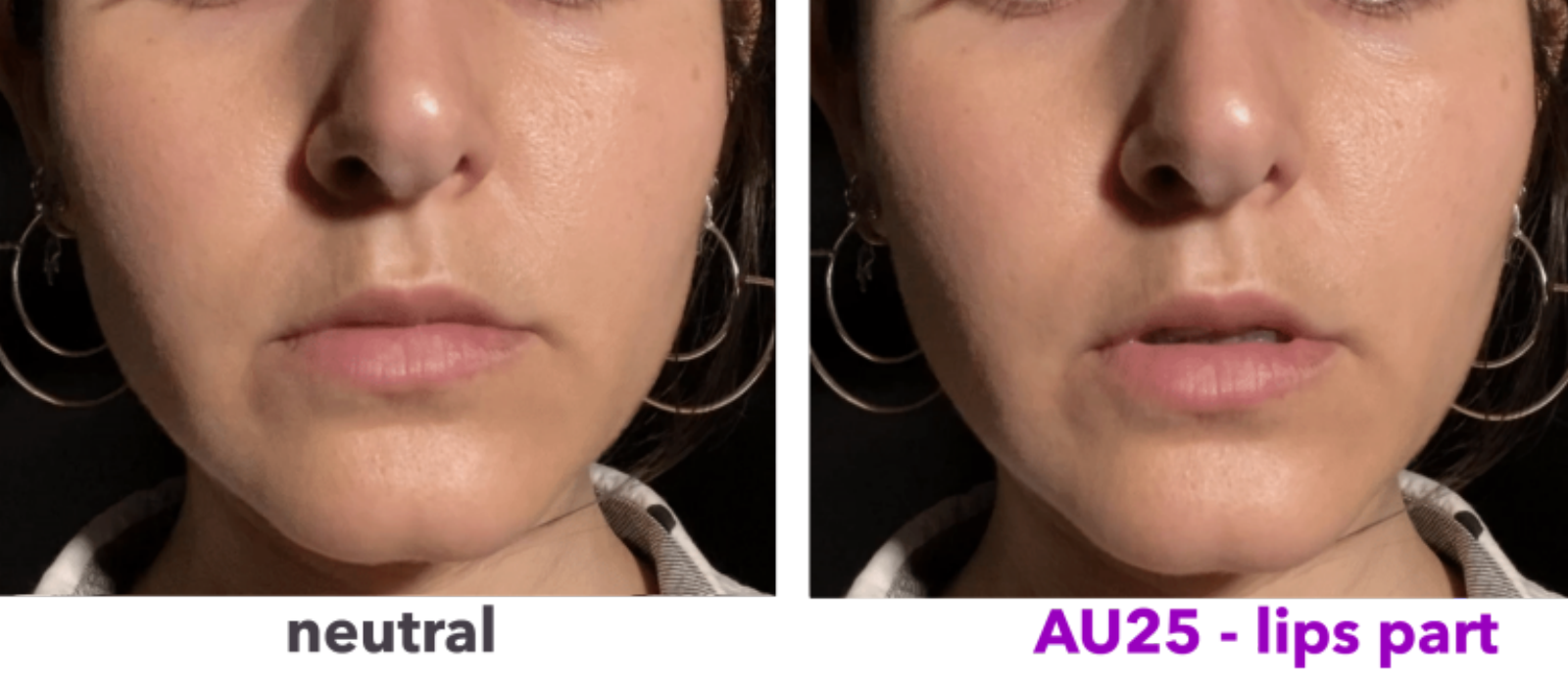}
  \end{subfigure}
  \hspace{0.08\linewidth}
  \begin{subfigure}{0.4\linewidth}
    \centering
    \includegraphics[width=\linewidth]{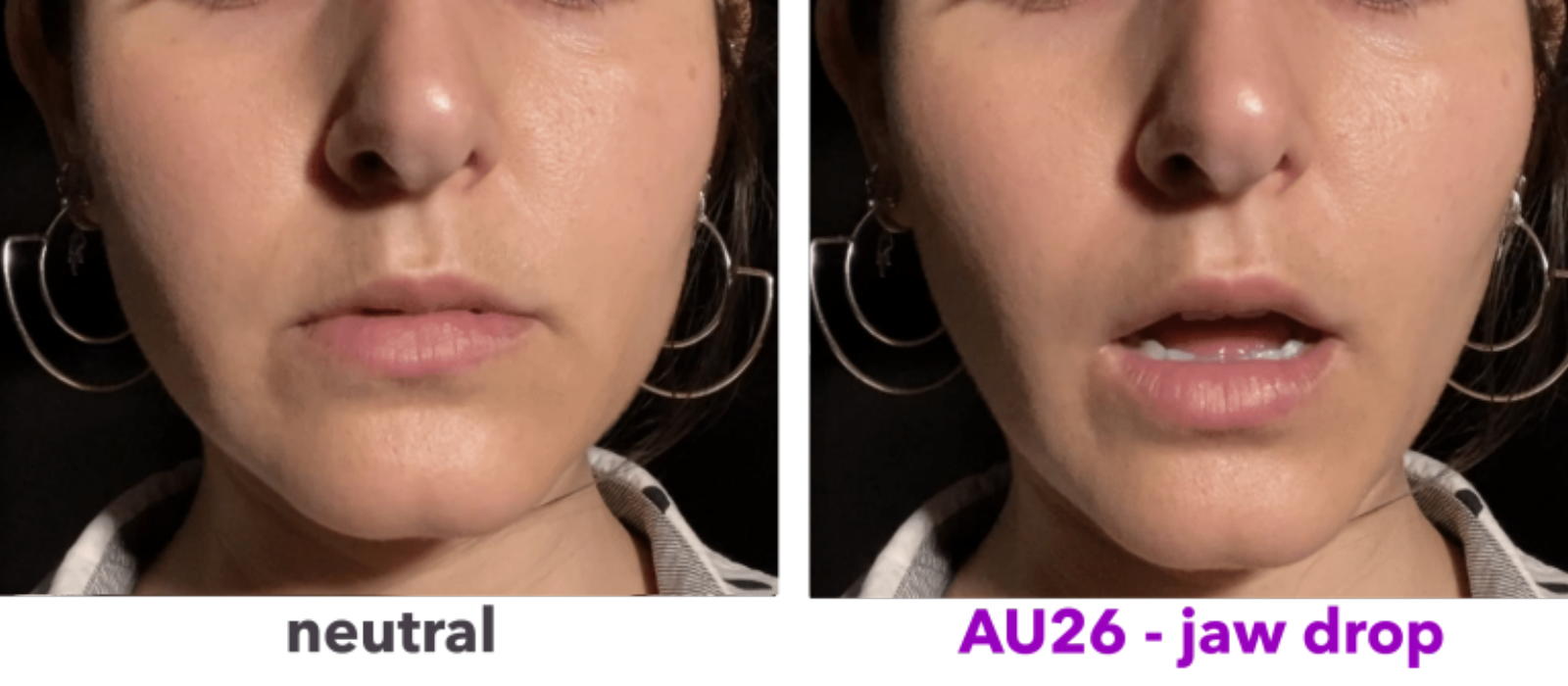}
  \end{subfigure}
  \caption{A visual reference guide for the primary AUs extracted from \cite{FACSlite}.}
  \label{fig:AUs}
\end{figure*}

\section{Visual Reference Guide for the Primary AUs}

See \Cref{fig:AUs} for a visual reference guide for the primary AUs analyzed in this paper (the AUs included in the DISFA and DISFA+ datasets).

\section{Equations for Loss Weight Computation}

We illustrated the method of computing the weights for the losses in our main paper. The exact equations for computing the weights are included in this section.

In AU intensity estimation, the weights for the MSE loss are defined as 
\begin{align}
  w_{i,j} = \begin{cases}
    \frac{2\cdot \frac{1}{\Sigma_{j'=0}^{1}n_{i,j'}}}{2\cdot\frac{1}{\Sigma_{j'=0}^{1}n_{i,j'}} + 4\cdot\frac{1}{\Sigma_{j'=2}^{5}n_{i,j'}}}, & \text{for $j=0,1$} \\
    \frac{4\cdot\frac{1}{\Sigma_{j'=2}^{5}n_{i,j'}}}{2\cdot\frac{1}{\Sigma_{j'=0}^{1}n_{i,j'}} + 4\cdot \frac{1}{\Sigma_{j'=2}^{5}n_{i,j'}}}, & \text{for $j=2,3,4,5$} \label{eqn:weights_reg}
  \end{cases}
\end{align}
while the weights for the cross entropy loss are defined as
\begin{align}
    \begin{cases}
    w_{i,j,1} = \frac{\frac{1}{\Sigma_{j'=j}^{5}n_{i,j'}}}{\Sigma_{j''=1}^{5}(\frac{1}{\Sigma_{j'=0}^{j''-1}n_{i,j'}} + \frac{1}{\Sigma_{j'=j''}^{5}n_{i,j'}})} \\
    w_{i,j,0} = \frac{\frac{1}{\Sigma_{j'=0}^{j-1}n_{i,j'}}}{\Sigma_{j''=1}^{5}(\frac{1}{\Sigma_{j'=0}^{j''-1}n_{i,j'}} + \frac{1}{\Sigma_{j'=j''}^{5}n_{i,j'}})},
    \label{eqn:weights_class}
    \end{cases}
\end{align}
where $n_{i,j}$ represents the number of occurrences of the $i$th AU with an intensity of $j$.

In AU detection, the weights for the cross entropy loss are defined as
\begin{align}
    \begin{cases}
    w_{i,1} = \frac{\frac{1}{n_{y_{i}\geq 2}}}{\frac{1}{n_{y_{i}\geq 2}}+\frac{1}{n_{y_{i} < 2}}} \\
    w_{i,0} = \frac{\frac{1}{n_{y_{i} < 2}}}{\frac{1}{n_{y_{i}\geq 2}}+\frac{1}{n_{y_{i} < 2}}},
    \label{eqn:weights_AU_detection}
    \end{cases}
\end{align}
where $n_{y_{i}\geq 2}$ and $n_{y_{i} < 2}$ represent the number of occurrences for $y_{i}\geq 2$ and $y_{i} < 2$ respectively.

\section{Full Results}

\Cref{tab:AU_intensity_estimation_performances_on_DISFA+,tab:AU_detection_performances_on_DISFA+,tab:AU_intensity_estimation_performances_on_UNBC-McMaster,tab:AU_detection_performances_on_UNBC-McMaster} present the breakdown of the results in \Cref{tab:AU_recognition_performances_on_DISFA+_and_UNBC-McMaster} for each AU. \Cref{tab:across_vs_within_participant_ICC_on_DISFA,tab:across_vs_within_participant_ICC_on_DISFA+,tab:across_vs_within_participant_ICC_on_UNBC-McMaster} present the breakdown of the results in \Cref{tab:across_vs_within_participant_ICC} for each AU. \Cref{tab:different_versions_of_CSN-IR50_on_AU_intensity_estimation,tab:different_versions_of_CSN-IR50_on_AU_detection} present the breakdown of the results in \Cref{tab:different_versions_of_CSN-IR50} for each AU.

\begin{table*}[t]
    \centering
    \begin{tabular}{|c|cccccccccccccc|}
        \hline
        \multirow{2}{*}{Metric} & \multirow{2}{*}{Method} & \multicolumn{12}{c}{AU} & \multirow{2}{*}{Average}\\
        \cline{3-14}

        & & 1 & 2 & 4 & 5 & 6 & 9 & 12 & 15 & 17 & 20 & 25 & 26 & \\ \hline
        \multirow{3}{*}{ICC(3,1)$\uparrow$} & IR50 (NCG) & .81 & .76 & \underline{.88} & .79 & .87 & .87 & .90 & .80 & .73 & \underline{.60} & .91 & .82 & .81\\
        & IR50 (OFC w/ BS) & .83 & .85 & .87 & .83 & .88 & .89 & .90 & \underline{.81} & .75 & \underline{.60} & \underline{.94} & .85 & .83\\
        & CSN-IR50 (OFC) & \underline{.90} & \underline{.92} & \underline{.88} & \underline{.89} & \underline{.90} & \underline{.91} & \underline{.92} & \underline{.81} & \underline{.80} & .57 & \underline{.94} & \underline{.88} & \underline{.86} \\
        \hline
        
        \multirow{3}{*}{MAE$\downarrow$} & IR50 (NCG) &.48 & .51 & .40 & .44 & .29 & .24 & .32 & .24 & .42 & .30 & .40 & .39 & .37\\
        & IR50 (OFC w/ BS) & .49 & .42 & .35 & .42 & .26 & .21 & .29 & .21 & .30 & .26 & .29 & .29 & .32\\
        & CSN-IR50 (OFC) & \underline{.26} & \underline{.22} & \underline{.29} & \underline{.29} & \underline{.23} & \underline{.18} & \underline{.24} & \underline{.17} & \underline{.21} & \underline{.21} & \underline{.24} & \underline{.24} & \underline{.23} \\
        \hline

    \end{tabular}
    \caption{The performance of different methods on AU intensity estimation on the DISFA+ dataset. For each metric, the best results in each column are underlined.}
    \label{tab:AU_intensity_estimation_performances_on_DISFA+}
\end{table*}

\begin{table*}[t]
    \centering
    \begin{tabular}{|c|cccccccccccccc|}
        \hline
        \multirow{2}{*}{Metric} & \multirow{2}{*}{Method} & \multicolumn{12}{c}{AU} & \multirow{2}{*}{Average}\\
        \cline{3-14}

        & & 1 & 2 & 4 & 5 & 6 & 9 & 12 & 15 & 17 & 20 & 25 & 26 & \\ \hline
        \multirow{3}{*}{F1 score$\uparrow$} & IR50 (NCG) & 71.0 & 63.1 & 84.0 & 63.7 & 85.9 & 51.0 & 84.5 & 64.7 & 58.8 & \underline{49.6} & 74.3 & 56.9 & 67.3\\
        & IR50 (OFC w/ BS) & 37.4 & 33.4 & 38.7 & 35.3 & 29.6 & 12.9 & 37.3 & 13.2 & 18.9 & 13.7 & 41.3 & 27.9 & 28.3\\
        & CSN-IR50 (OFC) & \underline{86.8} & \underline{84.0} & \underline{87.0} & \underline{79.5} & \underline{86.0} & \underline{86.4} & \underline{89.2} & \underline{70.1} & \underline{73.3} & 35.5 & \underline{95.0} & \underline{70.9} & \underline{78.6} \\
        \hline
        
        \multirow{3}{*}{Accuracy$\uparrow$} & IR50 (NCG) & 90.3 & 88.6 & 94.3 & 87.8 & 96.0 & 91.5 & 95.7 & 95.5 & 90.9 & 93.1 & 89.1 & 87.4 & 91.7\\
        & IR50 (OFC w/ BS) & 58.1 & 58.1 & 49.1 & 55.6 & 37.9 & 28.9 & 57.0 & 44.4 & 39.4 & 34.4 & 54.2 & 51.8 & 47.4\\
        & CSN-IR50 (OFC) & \underline{96.5} & \underline{96.3} & \underline{95.8} & \underline{94.6} & \underline{96.1} & \underline{98.6} & \underline{97.1} & \underline{97.4} & \underline{96.3} & \underline{93.9} & \underline{98.4} & \underline{93.8} & \underline{96.2} \\
        \hline

    \end{tabular}
    \caption{The performance of different methods on AU detection on the DISFA+ dataset. For each metric, the best results in each column are underlined.}
    \label{tab:AU_detection_performances_on_DISFA+}
\end{table*}

\begin{table*}[t]
    \centering
    \begin{tabular}{|c|ccccccccccc|}
        \hline
        \multirow{2}{*}{Metric} & \multirow{2}{*}{Method} & \multicolumn{9}{c}{AU} & \multirow{2}{*}{Average}\\
        \cline{3-11}

        & & 4 & 6 & 7 & 9 & 10 & 12 & 20 & 25 & 26 & \\ \hline
        \multirow{3}{*}{ICC(3,1)$\uparrow$} & IR50 (NCG) & .19 & .51 & .30 & .32 & .29 & .55 & \underline{.17} & .28 & .09 & .30\\
        & IR50 (OFC w/ BS) & .24 & .59 & .37 & .32 & .33 & .62 & \underline{.17} & .31 & .15 & .34\\
        & CSN-IR50 (OFC) & \underline{.41} & \underline{.67} & \underline{.49} & \underline{.47} & \underline{.49} & \underline{.70} & .14 & \underline{.39} & \underline{.29} & \underline{.45} \\
        \hline
        
        \multirow{3}{*}{MAE$\downarrow$} & IR50 (NCG) & .21 & .48 & .34 & .12 & .11 & .48 & .19 & .33 & .34 & .29\\
        & IR50 (OFC w/ BS) & .16 & .37 & .28 & .12 & .10 & .40 & .13 & \underline{.26} & .28 & .23\\
        & CSN-IR50 (OFC) & \underline{.12} & \underline{.28} & \underline{.27} & \underline{.08} & \underline{.08} & \underline{.37} & \underline{.11} & \underline{.26} & \underline{.22} & \underline{.20} \\
        \hline

    \end{tabular}
    \caption{The performance of different methods on AU intensity estimation on the UNBC-McMaster dataset. For each metric, the best results in each column are underlined.}
    \label{tab:AU_intensity_estimation_performances_on_UNBC-McMaster}
\end{table*}

\begin{table*}[t]
    \centering
    \begin{tabular}{|c|cccccccccccc|}
        \hline
        \multirow{2}{*}{Metric} & \multirow{2}{*}{Method} & \multicolumn{10}{c}{AU} & \multirow{2}{*}{Average}\\
        \cline{3-12}

        & & 4 & 6 & 7 & 9 & 10 & 12 & 20 & 25 & 26 & 43 & \\ \hline
        \multirow{3}{*}{F1 score$\uparrow$} & IR50 (NCG) & 17.2 & 47.1 & \underline{39.8} & 19.2 & 15.4 & 48.6 & 2.9 & 20.0 & 16.3 & 32.3 & 25.9\\
        & IR50 (OFC w/ BS) & 4.0 & 20.6 & 11.4 & 1.9 & 2.1 & 25.8 & 2.8 & 9.3 & 8.7 & 5.1 & 9.2\\
        & CSN-IR50 (OFC) & \underline{36.4} & \underline{56.1} & 39.3 & \underline{22.9} & \underline{24.5} & \underline{62.1} & \underline{5.5} & \underline{39.1} & \underline{26.9} & \underline{29.0} & \underline{34.2} \\
        \hline
        
        \multirow{3}{*}{Accuracy$\uparrow$} & IR50 (NCG) & 96.5 & 90.1 & \underline{95.8} & 95.9 & 95.4 & 88.3 & 94.8 & 86.7 & 92.7 & 96.5 & 93.3\\
        & IR50 (OFC w/ BS) & 29.9 & 40.8 & 35.8 & 29.6 & 31.6 & 46.3 & 39.7 & 38.9 & 35.0 & 33.4 & 36.1\\
        & CSN-IR50 (OFC) & \underline{98.1} & \underline{92.9} & 95.5 & \underline{97.3} & \underline{97.8} & \underline{92.4} & \underline{97.0} & \underline{94.7} & \underline{96.1} & \underline{96.7} & \underline{95.9} \\
        \hline

    \end{tabular}
    \caption{The performance of different methods on AU detection on the UNBC-McMaster dataset. For each metric, the best results in each column are underlined.}
    \label{tab:AU_detection_performances_on_UNBC-McMaster}
\end{table*}

\begin{table*}[t]
    \centering
    \begin{tabular}{|c|cccccccccccccc|}
        \hline
        \multirow{2}{*}{Metric} & \multirow{2}{*}{Method} & \multicolumn{12}{c}{AU} & \multirow{2}{*}{Average}\\
        \cline{3-14}

        & & 1 & 2 & 4 & 5 & 6 & 9 & 12 & 15 & 17 & 20 & 25 & 26 & \\ \hline
        \multirow{2}{*}{Across-Participant ICC(3,1)$\uparrow$} & IR50 (NCG) & .53 & .45 & .75 & .62 & .55 & .57 & .84 & \underline{.42} & .47 & .24 & .93 & .65 & .59\\
        & CSN-IR50 (OFC) & \underline{.75} & \underline{.70} & \underline{.80} & \underline{.72} & \underline{.67} & \underline{.61} & \underline{.85} & .33 & \underline{.52} & \underline{.37} & \underline{.94} & \underline{.77} & \underline{.67} \\
        \hline
        
        \multirow{2}{*}{Within-Participant ICC(3,1)$\uparrow$} & IR50 (NCG) & .40 & .35 & .66 & .38 & .54 & .46 & .83 & \underline{.27} & \underline{.45} & .26 & \underline{.93} & .57 & .51\\
        & CSN-IR50 (OFC) & \underline{.46} & \underline{.43} & \underline{.70} & \underline{.39} & \underline{.57} & \underline{.48} & \underline{.85} & .23 & .44 & \underline{.27} & \underline{.93} & \underline{.66} & \underline{.53} \\
        \hline

    \end{tabular}
    \caption{Comparison of within-participant ICC(3,1) averaged across all participants and across-participant ICC(3,1) between different methods on AU intensity estimation on the DISFA dataset. For each metric, the better results in each column are underlined.}
    \label{tab:across_vs_within_participant_ICC_on_DISFA}
\end{table*}

\begin{table*}[t]
    \centering
    \begin{tabular}{|c|cccccccccccccc|}
        \hline
        \multirow{2}{*}{Metric} & \multirow{2}{*}{Method} & \multicolumn{12}{c}{AU} & \multirow{2}{*}{Average}\\
        \cline{3-14}

        & & 1 & 2 & 4 & 5 & 6 & 9 & 12 & 15 & 17 & 20 & 25 & 26 & \\ \hline
        \multirow{2}{*}{Across-Participant ICC(3,1)$\uparrow$} & IR50 (NCG) & .81 & .76 & \underline{.88} & .79 & .87 & .87 & .90 & .80 & .73 & \underline{.60} & .91 & .82 & .81\\
        & CSN-IR50 (OFC) & \underline{.90} & \underline{.92} & \underline{.88} & \underline{.89} & \underline{.90} & \underline{.91} & \underline{.92} & \underline{.81} & \underline{.80} & .57 & \underline{.94} & \underline{.88} & \underline{.86} \\
        \hline
        
        \multirow{2}{*}{Within-Participant ICC(3,1)$\uparrow$} & IR50 (NCG) & .84 & .85 & \underline{.89} & .80 & .87 & .89 & .90 & \underline{.79} & .80 & \underline{.61} & \underline{.94} & .85 & .84\\
        & CSN-IR50 (OFC) & \underline{.89} & \underline{.91} & .88 & \underline{.86} & \underline{.90} & \underline{.91} & \underline{.92} & .76 & \underline{.82} & .57 & \underline{.94} & \underline{.88} & \underline{.85} \\
        \hline

    \end{tabular}
    \caption{Comparison of within-participant ICC(3,1) averaged across all participants and across-participant ICC(3,1) between different methods on AU intensity estimation on the DISFA+ dataset. For each metric, the better results in each column are underlined.}
    \label{tab:across_vs_within_participant_ICC_on_DISFA+}
\end{table*}

\begin{table*}[t]
    \centering
    \begin{tabular}{|c|ccccccccccc|}
        \hline
        \multirow{2}{*}{Metric} & \multirow{2}{*}{Method} & \multicolumn{9}{c}{AU} & \multirow{2}{*}{Average}\\
        \cline{3-11}

        & & 4 & 6 & 7 & 9 & 10 & 12 & 20 & 25 & 26 & \\ \hline
        \multirow{2}{*}{Across-Participant ICC(3,1)$\uparrow$} & IR50 (NCG) & .19 & .51 & .30 & .32 & .29 & .55 & \underline{.17} & .28 & .09 & .30\\
        & CSN-IR50 (OFC) & \underline{.41} & \underline{.67} & \underline{.49} & \underline{.47} & \underline{.49} & \underline{.70} & .14 & \underline{.39} & \underline{.29} & \underline{.45} \\
        \hline
        
        \multirow{2}{*}{Within-Participant ICC(3,1)$\uparrow$} & IR50 (NCG) & .11 & .53 & .24 & .18 & .05 & .50 & \underline{.07} & .22 & .11 & .22\\
        & CSN-IR50 (OFC) & \underline{.18} & \underline{.56} & \underline{.26} & \underline{.20} & \underline{.08} & \underline{.56} & \underline{.07} & \underline{.27} & \underline{.13} & \underline{.26} \\
        \hline

    \end{tabular}
    \caption{Comparison of within-participant ICC(3,1) averaged across all participants and across-participant ICC(3,1) between different methods on AU intensity estimation on the UNBC-McMaster dataset. For each metric, the better results in each column are underlined.}
    \label{tab:across_vs_within_participant_ICC_on_UNBC-McMaster}
\end{table*}

\begin{table*}[t]
    \centering
    \begin{tabular}{|c|cccccccccccccc|}
        \hline
        \multirow{2}{*}{Metric} & \multirow{2}{*}{Method} & \multicolumn{12}{c}{AU} & \multirow{2}{*}{Average}\\
        \cline{3-14}

        & & 1 & 2 & 4 & 5 & 6 & 9 & 12 & 15 & 17 & 20 & 25 & 26 & \\ \hline
        \multirow{6}{*}{ICC(3,1)$\uparrow$} & CSN-IR50-Stage1 & .61 & .57 & .76 & .60 & .66 & \underline{.61} & .84 & .26 & .48 & .32 & .92 & .71 & .61\\
        & CSN-IR50-Stage2 & .66 & .69 & .77 & .59 & .60 & .58 & .83 & .32 & .51 & .32 & .94 & \underline{.77} & .63\\
        & CSN-IR50-Stage3 & .68 & .68 & \underline{.81} & .67 & .64 & .59 & \underline{.85} & \underline{.36} & \underline{.55} & \underline{.38} & \underline{.95} & .74 & .66 \\
        & CSN-IR50-Stage4 & \underline{.75} & \underline{.70} & .80 & \underline{.72} & \underline{.67} & \underline{.61} & \underline{.85} & .33 & .52 & .37& .94 & \underline{.77} & \underline{.67} \\
        & CSN-IR50-FC & .57 & .50 & .71 & .51 & .57 & .55 & .79 & .29 & .40 & .19 & .88 & .61 & .55\\
        & CSN-IR50-Output & .58 & .49 & .70 & .48 & .55 & .54 & .79 & .35 & .41 & .21 & .87 & .56 & .54\\
        \hline
        
        \multirow{6}{*}{MAE$\downarrow$} & CSN-IR50-Stage1 & .34 & .31 & .48 & .13 & .31 & .23 & .36 & .24 & .35 & .23 & .33 & .41 & .31\\
        & CSN-IR50-Stage2 & .32 & .27 & .45 & .14 & .32 & .22 & .35 & .21 & .34 & .20 & .28 & .34 & .29\\
        & CSN-IR50-Stage3 & .28 & .24 & .42 & .11 & .32 & .22 & .33 & .19 & .26 & .18 & \underline{.25} & .32 & .26 \\
        & CSN-IR50-Stage4 & \underline{.19} & \underline{.16} & \underline{.38} & \underline{.08} & \underline{.26} & \underline{.19} & \underline{.31} & .17 & \underline{.22} & \underline{.13} & .27 & \underline{.27} & \underline{.22} \\
        & CSN-IR50-FC & .28 & .30 & .44 & .13 & .31 & .20 & .34 & .17 & .28 & .17 & .41 & .33 & .28\\
        & CSN-IR50-Output & .28 & .29 & .43 & .12 & .30 & .20 & .37 & \underline{.16} & .29 & .19 & .42 & .34 & .28\\
        \hline

    \end{tabular}
    \caption{The performance of different versions of CSN-IR50 on AU intensity estimation on the DISFA dataset.}
    \label{tab:different_versions_of_CSN-IR50_on_AU_intensity_estimation}
\end{table*}

\begin{table*}[t]
    \centering
    \begin{tabular}{|c|cccccccccc|}
        \hline
        \multirow{2}{*}{Metric} & \multirow{2}{*}{Method} & \multicolumn{8}{c}{AU} & \multirow{2}{*}{Average}\\
        \cline{3-10}

        & & 1 & 2 & 4 & 6 & 9 & 12 & 25 & 26 & \\ \hline
        \multirow{6}{*}{F1 score$\uparrow$} & CSN-IR50-Stage1 & 50.8 & 43.7 & 65.9 & 53.3 & 46.7 & 75.1 & 90.6 & 60.6 & 60.8\\
        & CSN-IR50-Stage2 & 54.0 & 54.7 & 70.5 & 52.3 & 48.4 & 77.2 & 93.4 & 71.5 & 65.2\\
        & CSN-IR50-Stage3 & 50.1 & 44.0 & \underline{73.5} & \underline{54.7} & 41.0 & 75.3 & 93.4 & \underline{67.3} & 62.4 \\
        & CSN-IR50-Stage4 & \underline{65.3} & \underline{58.3} & 70.8 & 52.6 & \underline{51.7} & \underline{77.3} & \underline{94.6} & 65.4 & \underline{67.0} \\
        & CSN-IR50-FC & 21.7 & 18.6 & 38.8 & 23.2 & 14.0 & 42.0 & 61.9 & 27.1 & 30.9\\
        & CSN-IR50-Output & 18.8 & 17.8 & 39.1 & 23.0 & 13.5 & 40.7 & 60.5 & 25.8 & 29.9\\
        \hline
        
        \multirow{6}{*}{Accuracy$\uparrow$} & CSN-IR50-Stage1 & 93.2 & 92.7 & 88.2 & 90.4 & 93.1 & 92.3 & 94.8 & 91.5 & 92.0\\
        & CSN-IR50-Stage2 & 93.4 & 95.0 & 89.9 & 90.5 & 94.1 & 93.3 & 96.3 & \underline{94.4} & 93.3\\
        & CSN-IR50-Stage3 & 92.3 & 93.7 & \underline{90.7} & 90.2 & 90.6 & 92.4 & 96.3 & 93.4 & 92.5 \\
        & CSN-IR50-Stage4 & \underline{96.9} & \underline{96.9} & 90.4 & \underline{91.7} & \underline{94.6} & \underline{93.5} & \underline{96.9} & 92.8 & \underline{94.2} \\
        & CSN-IR50-FC & 69.1 & 75.3 & 52.4 & 49.0 & 50.3 & 64.7 & 65.9 & 55.5 & 60.3\\
        & CSN-IR50-Output & 62.6 & 74.4 & 53.3 & 48.4 & 49.0 & 62.8 & 63.9 & 53.6 & 58.5\\
        \hline

    \end{tabular}
    \caption{The performance of different versions of CSN-IR50 on AU detection on the DISFA dataset.}
    \label{tab:different_versions_of_CSN-IR50_on_AU_detection}
\end{table*}

%% file: main.bib
@article{ekman1978facial,
  title={Facial action coding system},
  author={Ekman, Paul and Friesen, Wallace V},
  journal={Environmental Psychology \& Nonverbal Behavior},
  year={1978}
}

@article{mavadati2013disfa,
  title={{DISFA:} A spontaneous facial action intensity database},
  author={Mavadati, S Mohammad and Mahoor, Mohammad H and Bartlett, Kevin and Trinh, Philip and Cohn, Jeffrey F},
  journal={IEEE Transactions on Affective Computing},
  volume={4},
  number={2},
  pages={151--160},
  year={2013},
  publisher={IEEE}
}

@inproceedings{mavadati2016extended,
  title={Extended {DISFA} dataset: Investigating posed and spontaneous facial expressions},
  author={Mavadati, Mohammad and Sanger, Peyten and Mahoor, Mohammad H},
  booktitle={proceedings of the IEEE conference on computer vision and pattern recognition workshops},
  pages={1--8},
  year={2016}
}

@article{lugaresi2019mediapipe,
  title={{MediaPipe:} A framework for building perception pipelines},
  author={Lugaresi, Camillo and Tang, Jiuqiang and Nash, Hadon and McClanahan, Chris and Uboweja, Esha and Hays, Michael and Zhang, Fan and Chang, Chuo-Ling and Yong, Ming Guang and Lee, Juhyun and others},
  journal={arXiv preprint arXiv:1906.08172},
  year={2019}
}

@misc{PFL,
  title={{PyTorch Face Landmark}: A Fast and Accurate Facial Landmark Detector},
  url={https://github.com/cunjian/pytorch\face\_landmark},
  note={Open-source software available at https://github.com/cunjian/pytorch\_face\_landmark},
  author={Cunjian Chen},
  year={2021},
}

@inproceedings{kuo2018compact,
  title={A compact deep learning model for robust facial expression recognition},
  author={Kuo, Chieh-Ming and Lai, Shang-Hong and Sarkis, Michel},
  booktitle={Proceedings of the IEEE conference on computer vision and pattern recognition workshops},
  pages={2121--2129},
  year={2018}
}

@inproceedings{deng2019arcface,
  title={Arcface: Additive angular margin loss for deep face recognition},
  author={Deng, Jiankang and Guo, Jia and Xue, Niannan and Zafeiriou, Stefanos},
  booktitle={Proceedings of the IEEE/CVF conference on computer vision and pattern recognition},
  pages={4690--4699},
  year={2019}
}

@inproceedings{an2022killing,
  title={Killing two birds with one stone: Efficient and robust training of face recognition {CNN}s by partial {FC}},
  author={An, Xiang and Deng, Jiankang and Guo, Jia and Feng, Ziyong and Zhu, XuHan and Yang, Jing and Liu, Tongliang},
  booktitle={Proceedings of the IEEE/CVF Conference on Computer Vision and Pattern Recognition},
  pages={4042--4051},
  year={2022}
}

@inproceedings{niu2016ordinal,
  title={Ordinal regression with multiple output {CNN} for age estimation},
  author={Niu, Zhenxing and Zhou, Mo and Wang, Le and Gao, Xinbo and Hua, Gang},
  booktitle={Proceedings of the IEEE conference on computer vision and pattern recognition},
  pages={4920--4928},
  year={2016}
}

@article{shao2019facial,
  title={Facial action unit detection using attention and relation learning},
  author={Shao, Zhiwen and Liu, Zhilei and Cai, Jianfei and Wu, Yunsheng and Ma, Lizhuang},
  journal={IEEE transactions on affective computing},
  volume={13},
  number={3},
  pages={1274--1289},
  year={2019},
  publisher={IEEE}
}

@inproceedings{zhao2016deep,
  title={Deep region and multi-label learning for facial action unit detection},
  author={Zhao, Kaili and Chu, Wen-Sheng and Zhang, Honggang},
  booktitle={Proceedings of the IEEE conference on computer vision and pattern recognition},
  pages={3391--3399},
  year={2016}
}

@inproceedings{zhao2015joint,
  title={Joint patch and multi-label learning for facial action unit detection},
  author={Zhao, Kaili and Chu, Wen-Sheng and De la Torre, Fernando and Cohn, Jeffrey F and Zhang, Honggang},
  booktitle={Proceedings of the IEEE Conference on Computer Vision and Pattern Recognition},
  pages={2207--2216},
  year={2015}
}

@article{li2018eac,
  title={{EAC-Net}: Deep nets with enhancing and cropping for facial action unit detection},
  author={Li, Wei and Abtahi, Farnaz and Zhu, Zhigang and Yin, Lijun},
  journal={IEEE transactions on pattern analysis and machine intelligence},
  volume={40},
  number={11},
  pages={2583--2596},
  year={2018},
  publisher={IEEE}
}

@inproceedings{walecki2017deep,
  title={Deep structured learning for facial action unit intensity estimation},
  author={Walecki, Robert and Pavlovic, Vladimir and Schuller, Bj{\"o}rn and Pantic, Maja and others},
  booktitle={Proceedings of the IEEE Conference on Computer Vision and Pattern Recognition},
  pages={3405--3414},
  year={2017}
}

@inproceedings{XuNIPS2019,
author="Xiaoxing Xu and J.S. Huang and V.R. de Sa",
title="Pain Evaluation in Video using Extended Multitask Learning from Multidimensional Measurements",
booktitle="Proceedings of Machine Learning Research, (Machine Learning for Health ML4H at NeurIPS 2019)",
year=2019}

@inproceedings{mavadati2012automatic,
  title={Automatic detection of non-posed facial action units},
  author={Mavadati, S Mohammad and Mahoor, Mohammad H and Bartlett, Kevin and Trinh, Philip},
  booktitle={2012 19th IEEE International Conference on Image Processing},
  pages={1817--1820},
  year={2012},
  organization={IEEE}
}

@inproceedings{baltruvsaitis2016openface,
  title={Openface: an open source facial behavior analysis toolkit},
  author={Baltru{\v{s}}aitis, Tadas and Robinson, Peter and Morency, Louis-Philippe},
  booktitle={2016 IEEE winter conference on applications of computer vision (WACV)},
  pages={1--10},
  year={2016},
  organization={IEEE}
}

@inproceedings{linh2017deepcoder,
  title={Deepcoder: Semi-parametric variational autoencoders for automatic facial action coding},
  author={Linh Tran, Dieu and Walecki, Robert and Eleftheriadis, Stefanos and Schuller, Bjorn and Pantic, Maja and others},
  booktitle={Proceedings of the IEEE International Conference on Computer Vision},
  pages={3190--3199},
  year={2017}
}

@inproceedings{fan2020facial,
  title={Facial action unit intensity estimation via semantic correspondence learning with dynamic graph convolution},
  author={Fan, Yingruo and Lam, Jacqueline and Li, Victor},
  booktitle={Proceedings of the AAAI Conference on Artificial Intelligence},
  volume={34},
  number={07},
  pages={12701--12708},
  year={2020}
}

@misc{FACSlite,
  title={{FACS LITE -- STILL \& FAST}},
  url={https://melindaozel.com/facs-lite-still-fast},
  note={A visual reference guide for FACS at https://melindaozel.com/facs-lite-still-fast (last accessed on 03/13/2024).},
  author={Melinda Ozel},
  year={2024},
}

@article{liu2023multi,
  title={Multi-scale promoted self-adjusting correlation learning for facial action unit detection},
  author={Liu, Xin and Yuan, Kaishen and Niu, Xuesong and Shi, Jingang and Yu, Zitong and Yue, Huanjing and Yang, Jingyu},
  journal={arXiv preprint arXiv:2308.07770},
  year={2023}
}

@article{yuan2024auformer,
  title={{AUformer}: Vision transformers are parameter-efficient facial action unit detectors},
  author={Yuan, Kaishen and Yu, Zitong and Liu, Xin and Xie, Weicheng and Yue, Huanjing and Yang, Jingyu},
  journal={arXiv preprint arXiv:2403.04697},
  year={2024}
}

@inproceedings{zhang2023weakly,
  title={Weakly-supervised text-driven contrastive learning for facial behavior understanding},
  author={Zhang, Xiang and Wang, Taoyue and Li, Xiaotian and Yang, Huiyuan and Yin, Lijun},
  booktitle={Proceedings of the IEEE/CVF International Conference on Computer Vision},
  pages={20751--20762},
  year={2023}
}

@inproceedings{song2021uncertain,
  title={Uncertain graph neural networks for facial action unit detection},
  author={Song, Tengfei and Chen, Lisha and Zheng, Wenming and Ji, Qiang},
  booktitle={Proceedings of the AAAI Conference on Artificial Intelligence},
  volume={35},
  number={7},
  pages={5993--6001},
  year={2021}
}

@article{shao2021jaa,
  title={{J\^AA-Net}: joint facial action unit detection and face alignment via adaptive attention},
  author={Shao, Zhiwen and Liu, Zhilei and Cai, Jianfei and Ma, Lizhuang},
  journal={International Journal of Computer Vision},
  volume={129},
  pages={321--340},
  year={2021},
  publisher={Springer}
}

@inproceedings{tang2021piap,
  title={{PIAP-DF}: Pixel-interested and anti person-specific facial action unit detection net with discrete feedback learning},
  author={Tang, Yang and Zeng, Wangding and Zhao, Dafei and Zhang, Honggang},
  booktitle={Proceedings of the IEEE/CVF International Conference on Computer Vision},
  pages={12899--12908},
  year={2021}
}

@article{luo2022learning,
  title={Learning multi-dimensional edge feature-based AU relation graph for facial action unit recognition},
  author={Luo, Cheng and Song, Siyang and Xie, Weicheng and Shen, Linlin and Gunes, Hatice},
  journal={arXiv preprint arXiv:2205.01782},
  year={2022}
}

@inproceedings{chang2022knowledge,
  title={Knowledge-driven self-supervised representation learning for facial action unit recognition},
  author={Chang, Yanan and Wang, Shangfei},
  booktitle={Proceedings of the IEEE/CVF Conference on Computer Vision and Pattern Recognition},
  pages={20417--20426},
  year={2022}
}

@inproceedings{wang2024multi,
  title={Multi-scale Dynamic and Hierarchical Relationship Modeling for Facial Action Units Recognition},
  author={Wang, Zihan and Song, Siyang and Luo, Cheng and Deng, Songhe and Xie, Weicheng and Shen, Linlin},
  booktitle={Proceedings of the IEEE/CVF Conference on Computer Vision and Pattern Recognition},
  pages={1270--1280},
  year={2024}
}

@book{ekman2002facial,
  title={Facial Action Coding System: The Manual},
  author={Ekman, Paul and Friesen, Wallace V and Hager, Joseph C},
  year={2002},
  publisher={A Human Face}
}

@inproceedings{shao2018deep,
  title={Deep adaptive attention for joint facial action unit detection and face alignment},
  author={Shao, Zhiwen and Liu, Zhilei and Cai, Jianfei and Ma, Lizhuang},
  booktitle={Proceedings of the European conference on computer vision (ECCV)},
  pages={705--720},
  year={2018}
}

@inproceedings{lucey2011painful,
  title={Painful data: The UNBC-McMaster shoulder pain expression archive database},
  author={Lucey, Patrick and Cohn, Jeffrey F and Prkachin, Kenneth M and Solomon, Patricia E and Matthews, Iain},
  booktitle={2011 IEEE International Conference on Automatic Face \& Gesture Recognition (FG)},
  pages={57--64},
  year={2011},
  organization={IEEE}
}

@article{churamani2022domain,
  title={Domain-incremental continual learning for mitigating bias in facial expression and action unit recognition},
  author={Churamani, Nikhil and Kara, Ozgur and Gunes, Hatice},
  journal={IEEE Transactions on Affective Computing},
  volume={14},
  number={4},
  pages={3191--3206},
  year={2022},
  publisher={IEEE}
}

@article{kara2021towards,
  title={Towards fair affective robotics: continual learning for mitigating bias in facial expression and action unit recognition},
  author={Kara, Ozgur and Churamani, Nikhil and Gunes, Hatice},
  journal={arXiv preprint arXiv:2103.09233},
  year={2021}
}

@article{domnich2021responsible,
  title={Responsible AI: Gender bias assessment in emotion recognition},
  author={Domnich, Artem and Anbarjafari, Gholamreza},
  journal={arXiv preprint arXiv:2103.11436},
  year={2021}
}

@article{sham2023ethical,
  title={Ethical AI in facial expression analysis: racial bias},
  author={Sham, Abdallah Hussein and Aktas, Kadir and Rizhinashvili, Davit and Kuklianov, Danila and Alisinanoglu, Fatih and Ofodile, Ikechukwu and Ozcinar, Cagri and Anbarjafari, Gholamreza},
  journal={Signal, Image and Video Processing},
  volume={17},
  number={2},
  pages={399--406},
  year={2023},
  publisher={Springer}
}

@inproceedings{kim2021age,
  title={Age bias in emotion detection: An analysis of facial emotion recognition performance on young, middle-aged, and older adults},
  author={Kim, Eugenia and Bryant, De'Aira and Srikanth, Deepak and Howard, Ayanna},
  booktitle={Proceedings of the 2021 AAAI/ACM Conference on AI, Ethics, and Society},
  pages={638--644},
  year={2021}
}

@inproceedings{chen2021understanding,
  title={Understanding and mitigating annotation bias in facial expression recognition},
  author={Chen, Yunliang and Joo, Jungseock},
  booktitle={Proceedings of the IEEE/CVF International Conference on Computer Vision},
  pages={14980--14991},
  year={2021}
}

@article{zhang2014bp4d,
  title={{BP4D-spontaneous}: a high-resolution spontaneous 3d dynamic facial expression database},
  author={Zhang, Xing and Yin, Lijun and Cohn, Jeffrey F and Canavan, Shaun and Reale, Michael and Horowitz, Andy and Liu, Peng and Girard, Jeffrey M},
  journal={Image and Vision Computing},
  volume={32},
  number={10},
  pages={692--706},
  year={2014},
  publisher={Elsevier}
}

@inproceedings{suresh2022using,
  title={Using positive matching contrastive loss with facial action units to mitigate bias in facial expression recognition},
  author={Suresh, Varsha and Ong, Desmond C},
  booktitle={2022 10th International Conference on Affective Computing and Intelligent Interaction (ACII)},
  pages={1--8},
  year={2022},
  organization={IEEE}
}

@inproceedings{zhang2018identity,
  title={Identity-based Adversarial Training of Deep CNNs for Facial Action Unit Recognition.},
  author={Zhang, Zheng and Zhai, Shuangfei and Yin, Lijun and others},
  booktitle={BMVC},
  pages={226},
  year={2018},
  organization={Newcastle}
}

@inproceedings{tu2019idennet,
  title={Idennet: Identity-aware facial action unit detection},
  author={Tu, Cheng-Hao and Yang, Chih-Yuan and Hsu, Jane Yung-jen},
  booktitle={2019 14th IEEE International Conference on Automatic Face \& Gesture Recognition (FG 2019)},
  pages={1--8},
  year={2019},
  organization={IEEE}
}

@inproceedings{meng2017identity,
  title={Identity-aware convolutional neural network for facial expression recognition},
  author={Meng, Zibo and Liu, Ping and Cai, Jie and Han, Shizhong and Tong, Yan},
  booktitle={2017 12th IEEE International Conference on Automatic Face \& Gesture Recognition (FG 2017)},
  pages={558--565},
  year={2017},
  organization={IEEE}
}

@article{zhang2020identity,
  title={Identity--expression dual branch network for facial expression recognition},
  author={Zhang, Haifeng and Su, Wen and Yu, Jun and Wang, Zengfu},
  journal={IEEE transactions on cognitive and developmental systems},
  volume={13},
  number={4},
  pages={898--911},
  year={2020},
  publisher={IEEE}
}

@article{bromley1993signature,
  title={Signature verification using a {``Siamese''} time delay neural network},
  author={Bromley, Jane and Guyon, Isabelle and LeCun, Yann and S{\"a}ckinger, Eduard and Shah, Roopak},
  journal={Advances in neural information processing systems},
  volume={6},
  year={1993}
}

@inproceedings{yang2017neural,
  title={Neural aggregation network for video face recognition},
  author={Yang, Jiaolong and Ren, Peiran and Zhang, Dongqing and Chen, Dong and Wen, Fang and Li, Hongdong and Hua, Gang},
  booktitle={Proceedings of the IEEE conference on computer vision and pattern recognition},
  pages={4362--4371},
  year={2017}
}

@inproceedings{taigman2014deepface,
  title={Deepface: Closing the gap to human-level performance in face verification},
  author={Taigman, Yaniv and Yang, Ming and Ranzato, Marc'Aurelio and Wolf, Lior},
  booktitle={Proceedings of the IEEE conference on computer vision and pattern recognition},
  pages={1701--1708},
  year={2014}
}

@inproceedings{Fabi22,
author="Sarah Fabi and Xiaojing Xu and Virginia R. de Sa",
title="Exploring the Racial Bias in Pain Detection with a Computer Vision Model",
booktitle="Proceedings of the 44th Annual Meeting of the Cognitive Science Society",
year = 2022,
editor="J. Culbertson and A. Perfors and H. Rabagliati and V Ramenzoni",
pages="358-365"}

@inproceedings{xu2021personalized,
  title={Personalized Pain Detection in Facial Video with Uncertainty Estimation},
  author={Xu, Xiaojing and de Sa, Virginia R},
  booktitle={{International Conference of the IEEE Engineering in Medicine \& Biology Society}},
  pages={4163--4168},
  year={2021},
  organization={IEEE}
}

@inproceedings{Raina22,
author="Ritik Raina and Miguel Monares and Mingze Xu and Sarah Fabi and Xiaojing Xu and Lehan Li and William Sumerfield and Jin Gan and Virginia R. de Sa",
title="Exploring Biases in Facial Expression Analysis using Synthetic Faces",
booktitle="Neural Information Processing Systems Workshop on Synthetic Data for Empowering ML Research (SyntheticData4ML)",
year=2022,
url="https://neurips.cc/virtual/2022/58648"}

@inproceedings{Monares23,
    author = "Miguel Monares and Yuan Tang and Ritik Raina and Virginia R. de Sa",
    title = "Analyzing Biases in AU Activation Estimation Toward Fairer Facial Expression Recognition",
    booktitle = "KDD '23",
    year = 2023}

@inproceedings{girard2017sayette,
  title={Sayette group formation task ({GFT}) spontaneous facial expression database},
  author={Girard, Jeffrey M and Chu, Wen-Sheng and Jeni, L{\'a}szl{\'o} A and Cohn, Jeffrey F},
  booktitle={2017 12th IEEE international conference on automatic face \& gesture recognition (FG 2017)},
  pages={581--588},
  year={2017},
  organization={IEEE}
}

@article{mcduff2018fed+,
  title={{AM-FED+}: An extended dataset of naturalistic facial expressions collected in everyday settings},
  author={McDuff, Daniel and Amr, May and El Kaliouby, Rana},
  journal={IEEE Transactions on Affective Computing},
  volume={10},
  number={1},
  pages={7--17},
  year={2018},
  publisher={IEEE}
}

@inproceedings{valstar2006fully,
  title={Fully automatic facial action unit detection and temporal analysis},
  author={Valstar, Michel and Pantic, Maja},
  booktitle={2006 Conference on Computer Vision and Pattern Recognition Workshop (CVPRW'06)},
  pages={149--149},
  year={2006},
  organization={IEEE}
}

@inproceedings{baltruvsaitis2015cross,
  title={Cross-dataset learning and person-specific normalisation for automatic action unit detection},
  author={Baltru{\v{s}}aitis, Tadas and Mahmoud, Marwa and Robinson, Peter},
  booktitle={2015 11th IEEE international conference and workshops on automatic face and gesture recognition (FG)},
  volume={6},
  pages={1--6},
  year={2015},
  organization={IEEE}
}

@inproceedings{yang2019learning,
  title={Learning temporal information from a single image for AU detection},
  author={Yang, Huiyuan and Yin, Lijun},
  booktitle={2019 14th IEEE International Conference on Automatic Face \& Gesture Recognition (FG 2019)},
  pages={1--8},
  year={2019},
  organization={IEEE}
}

@inproceedings{hernandez2022deepfn,
  title={{DeepFN}: towards generalizable facial action unit recognition with deep face normalization},
  author={Hernandez, Javier and McDuff, Daniel and Rudovic, Ognjen Oggi and Fung, Alberto and Czerwinski, Mary},
  booktitle={2022 10th International Conference on Affective Computing and Intelligent Interaction (ACII)},
  pages={1--8},
  year={2022},
  organization={IEEE}
}
